\documentclass[lettersize,journal]{IEEEtran}
\usepackage{algorithmic}
\usepackage{algorithm}
\usepackage{array}
\usepackage[caption=false,font=footnotesize,labelfont=rm,textfont=rm]{subfig}

\usepackage{textcomp}
\usepackage{stfloats}
\usepackage{url}
\usepackage{verbatim}
\usepackage{graphicx}
\usepackage{cite}
\usepackage{amsmath,amssymb,amsfonts}
\usepackage{bm} 
\usepackage{enumitem}
\usepackage{hyperref}
\usepackage{placeins}
\usepackage{booktabs}

\usepackage{xcolor}
\usepackage{color}

\begin{document}

\title{Semantic-Aware Joint Source-Channel Optimization for Encoder-Agnostic Digital Video Communication}

\author{Xiangben Zhu,~\IEEEmembership{Graduate Student Member,~IEEE,}
Caili Guo,~\IEEEmembership{Senior Member,~IEEE,}
Yang Yang,~\IEEEmembership{Senior Member,~IEEE,}
Chuanhong Liu,
and Meiyi Zhu
\thanks{This paper has been partially presented in IEEE WCNC 2026~\cite{zhu2026semantic}}
\thanks{\protect\raggedright Xiangben Zhu, Caili Guo and Yang Yang are with Beijing Key Laboratory of Network System Architecture and Convergence, School of Information and Communication Engineering, Beijing University of Posts and Telecommunications, Beijing 100876, China. (email: zhuxiangben13 @bupt.edu.cn;guocaili@bupt.edu.cn;
yangyang01@bupt.edu.cn)}
\thanks{Chuanhong Liu is with China Mobile (Suzhou) Software Technology Company Limited, Suzhou 215011, China. (email: liuchuan hong@cmss.chinamobile.com)}
\thanks{Meiyi Zhu is with the Department of Engineering, King's College London, London WC2R 2LS, U.K. (e-mail: meiyi.1.zhu@kcl.ac.uk)}
\thanks{ }
}

\markboth{Journal of \LaTeX\ Class Files}%
{zhu \MakeLowercase{\textit{et al.}}: SAJSCO}

\IEEEpubid{0000--0000/00\$00.00~\copyright~2026 IEEE}

\maketitle

\begin{abstract}
Video semantic communication has attracted increasing attention as a promising approach to improving video transmission efficiency. However, most existing approaches rely on computationally intensive deep learning-based video encoders and decoders, which hinders their deployment in resource-constrained scenarios. To address this issue, we propose a lightweight semantic-aware joint source-channel optimization (SAJSCO) scheme that can be integrated into existing digital video communication systems as a plug-in module.
Specifically, we develop a video communication system model in which the transmitter jointly optimizes source and channel coding parameters based on the inter-frame semantic importance of the input video and estimated channel state information.
On this basis, we formulate an optimization problem that maximizes semantic importance weighted video reconstruction quality under a maximum bitrate constraint.
To solve it, we first quantify inter-frame semantic importance using a cosine similarity-based metric with a shifted window mechanism. We then develop a multi-actor proximal policy optimization (MPPO) algorithm to solve the formulated problem by jointly adapting the source compression rate and channel coding rate. The learned policy can be directly applied to different video encoders without encoder-specific retraining or fine-tuning. 
SAJSCO achieves Bjøntegaard Delta rate reductions of 34.86\% and 18.01\% when integrated with H.265, a conventional video encoder, and DCVC-RT, a deep learning-based video encoder. Over-the-air experiments on a hardware testbed further demonstrate a PSNR gain of up to 1.448 dB with H.265 and an LPIPS reduction of up to 0.033 with DCVC-RT compared with the respective best-performing fixed-parameter baselines.
\end{abstract}

\begin{IEEEkeywords}
Video semantic communication, temporal video semantics, deep reinforcement learning, prototype validation.
\end{IEEEkeywords}

\section{Introduction}
\setlength{\parskip}{0pt}  

The deep integration of artificial intelligence (AI) and 6G communication is becoming an inevitable trend. The proliferation of multimedia devices has led to a surge in video traffic, while deep neural networks for video encoding require substantial computational resources, posing significant challenges to 6G wireless communication systems~\cite{sanjalawe2025review}. 
However, conventional video communication systems based on bit-level transmission and codecs such as H.265~\cite{sullivan2012overview} still suffer from limited compression efficiency and are vulnerable to channel impairments, where errors in critical control bits may cause decoding failures~\cite{li2026goal}. Video semantic communication (VSC), leveraging artificial intelligence technologies, enhances coding efficiency and improves communication robustness, thereby making it a key enabling technology for 6G networks~\cite{gao2025cross}.

Existing video semantic communication methods have demonstrated 
considerable success in reducing transmission bandwidth while 
maintaining perceptual quality. A representative line of research 
leverages deep neural networks to jointly optimize source and channel 
coding in an end-to-end manner~\cite{tung2022deepwive, wang2022wireless, liang2024generative}. However, such approaches impose heavy computational burdens on communication systems and often lack explicit adaptation to time-varying channel conditions, making practical deployment challenging. 
Another line of work incorporates semantic awareness into conventional codec-based pipelines~\cite{wang2023semantic}, offering greater compatibility with existing infrastructures, yet most of these methods focus on spatial semantics such as regions of interest~\cite{liang2023vista, aliouat2023region}. The temporal semantics inherent in video content, which capture motion patterns and inter-frame dependencies are pivotal for inter-frame prediction and redundancy elimination~\cite{tian2023non, zhang2023neural}, remain largely underexplored. Motivated by these observations, this work proposes a low-complexity video semantic communication scheme that exploits temporal semantic importance to guide the joint optimization of source and channel coding, while maintaining full compatibility with existing digital communication systems.

\subsection{Related Work}
\IEEEpubidadjcol

Existing approaches fall into two broad categories, namely deep neural network-based methods and conventional codec-based methods, which are detailed in the following.

\textit{1) Deep video semantic communication}: Deep video semantic communication employs deep neural networks as semantic encoders and decoders, encoding videos into compact semantic features to reduce transmission bandwidth~\cite{tung2022deepwive}. 
The use of deep learning-based encoders naturally offers advantages in security, as demonstrated in~\cite{he2024secure}, where adversarial residual networks are employed to enhance the privacy of semantic communication.
To achieve intelligent task-oriented communication, the authors in~\cite{yang2022semantic, liu2023adaptable} developed semantic communication systems that focus on transmitting only task-relevant semantic information, thereby reducing bandwidth consumption. Furthermore, the authors in~\cite{liu2025lightweight} introduced a lightweight task-oriented semantic communication framework empowered by large-scale AI models, where knowledge distillation is leveraged to reduce model complexity and computational latency.
The authors in~\cite{chen2024generative} utilized artificial intelligence generated content (AIGC) to transmit multi-modal semantics and generate videos at the receiver. However, most deep learning-based video encoders introduce considerable computational overhead, rendering real-time encoding of 1080p video at 25 fps unattainable even with state-of-the-art GPU acceleration, posing a fundamental barrier to practical deployment~\cite{jia2025towards}.

\textit{2) Semantic-driven video communication}: Semantic-driven video communication regards semantic information as prior knowledge to optimize existing video communication frameworks.
The authors in~\cite{liu2024ofdm} allocated communication resources based on semantic importance, thereby improving task performance. The authors in~\cite{liang2023vista} separated dynamic and static contents of videos, and encoded their semantics and locations.
Spatial semantic importance has been incorporated into ROI-based video coding and resource allocation to prioritize high-quality transmission of semantically important regions~\cite{wang2023semantic, aliouat2023region}. However, existing ROI-based semantic communication methods mainly focus on intra-frame semantic importance while ignoring inter-frame semantics, leading to residual temporal redundancy. Deep reinforcement learning has been explored for video communication by modeling adaptive coding and streaming optimization as sequential decision-making problems, enabling dynamic parameter adjustment under varying video content and network conditions~\cite{huang2021adaptive,li2023toward,zhou2020rate}. Temporal redundancy across video frames can be exploited to reduce redundant bits and improve bandwidth efficiency, while temporally important semantic content should be assigned more transmission resources for reliable reconstruction~\cite{zhang2023neural}. 
Therefore, the exploitation of temporal semantics for video reconstruction remains insufficiently explored.

\vspace{-0.5em}
\subsection{Challenges and Contributions}
Deep-learning based semantic encoders impose substantial computational overhead. Recently, a growing number of studies~\cite{yin2025generative, qiao2025token} have investigated the use of generative large models for video semantic communication, in which only text-level prompts and key frames are transmitted. However, this paradigm entails substantial decoding complexity and imposes considerable computational demands at the receiver. Although considerable efforts have been devoted to lightweight model design~\cite{peng2025semantic}, their AI computing requirements remain high, making them difficult to deploy in existing communication systems. Fortunately, this issue can be alleviated by integrating semantic communication with existing communication systems, where the video is first semantically understood and then encoded using conventional video encoders.

In addition, video is a form of spatiotemporal data characterized by temporal evolution and two-dimensional spatial pixels, and its semantic importance is inherently non-uniform. Such semantic importance can provide valuable guidance for video encoding. The study in~\cite{wang2023semantic, gao2025cross} investigated the semantic importance of video in the spatial dimension and allocated communication resources to different regions according to their semantic importance, thereby improving coding efficiency. To more effectively compress redundant video information and ensure robust transmission, it is essential to investigate the semantic importance of video in the temporal dimension. Moreover, due to the fluctuations of the signal-to-noise ratio (SNR) in real-world video transmission scenarios, it is necessary to jointly consider video semantics and channel state information to design a robust and efficient video transmission mechanism. The work in~\cite{yang2024swinjscc} exploits instantaneous channel state information (CSI) to adaptively rescale image semantic features, thereby enabling implicit joint source–channel rate adaptation. In the study of performance optimization for semantic-aware video transmission, several challenges must be addressed:

\textit{Challenge 1: How to enable real-time video semantic communication under the computational constraints of existing communication infrastructure?}

\textit{Challenge 2: Under computational constraints, how to effectively measure the temporal, or inter-frame, semantic importance of video?}

\textit{Challenge 3:  How to leverage the measured temporal semantic importance together with real-time CSI to achieve effective joint source-channel optimization?}

In this paper, we first develop a semantic-aware joint source-channel optimization (SAJSCO) scheme for video transmission. We then propose a novel method for measuring semantic importance and devise a deep reinforcement learning-based algorithm to determine the optimal source coding and channel coding parameters. A preliminary version of the work on video source coding optimization was introduced in our conference paper. To the best of our knowledge, \textit{this is the first work to investigate video coding parameter optimization from the perspective of inter-frame semantic importance}. The main contributions of this paper are summarized as follows:
\begin{enumerate}[label=\arabic*), topsep=0pt, partopsep=0pt, itemsep=0pt, parsep=0pt]
    \item We propose a joint source–channel optimization framework that employs lightweight video semantic extraction and analysis. This framework has low computational requirements and can seamlessly integrate with existing digital video communication systems, offering good robustness against fluctuating channel conditions. Considering the varying inter-frame semantic importance of the video, we formulate an optimization problem for video rate-distortion, with the objective of optimizing the semantic importance weighted video reconstruction quality under the constraint of maximum bitrate. This addresses the aforementioned \textit{Challenge 1}.

    \item To solve the problem, We propose an inter-frame semantic importance evaluation method that models temporal correlations based on cosine similarity in the semantic feature domain. By incorporating a shifted window mechanism, the method captures local inter-GOP dependencies and aggregates them progressively, yielding a fine-grained importance representation that effectively guides parameter selection. This addresses the aforementioned \textit{Challenge 2}.

    \item Then, we propose a deep reinforcement learning-based parameter selection algorithm. Specifically, the algorithm comprehensively considers both the inter-frame semantic importance of the video and real-time CSI feedback, achieving a balance between efficient video compression and reliable transmission. Furthermore, a multi-actor reinforcement learning algorithm is designed to separately select the compression rate for source encoding and the coding rate for channel encoding. The multi-actor approach enables the network to learn the characteristics and strategies for both source and channel encoding parameters, thereby enhancing overall performance optimization. This effectively addresses the aforementioned \textit{Challenge 3}.

    \item Experimental results demonstrate that the proposed deep reinforcement learning-based scheme outperforms both traditional and AI-based video communication methods. Additionally, the proposed parameter selection algorithm yields substantial improvements in video quality, achieving better results under both fixed Signal-to-Noise Ratio (SNR) and actual fluctuating SNR conditions.
    
\end{enumerate}

The remainder of this paper is organized as follows. Sec.~\ref{sec:SYSTEM MODEL AND PROBLEM FORMULATION} introduces the system model and problem formulation. Sec.~\ref{sec:SEMANTIC IMPORTANCE EVALUATION} introduces the proposed video inter-frame semantic importance measurement method. The proposed deep reinforcement learning algorithm is detailed in Sec.~\ref{sec:PROPOSED ALGORITHM}. Sec.~\ref{sec:SIMULATION RESULTS AND ANALYSIS} presents the analysis of simulation results. Sec.~\ref{sec:PROTOTYPE VALIDATION} presents the hardware prototype implementation and experimental validation conducted on a physical platform to verify the practical feasibility of the proposed system. Finally, Sec.~\ref{sec:CONCLUSION} summarizes the key conclusions of this study.

\section{System model and problem formulation}\label{sec:SYSTEM MODEL AND PROBLEM FORMULATION}

\begin{figure*}[t]
    \centering
    \includegraphics[width=\linewidth, height=0.26\textheight, keepaspectratio]{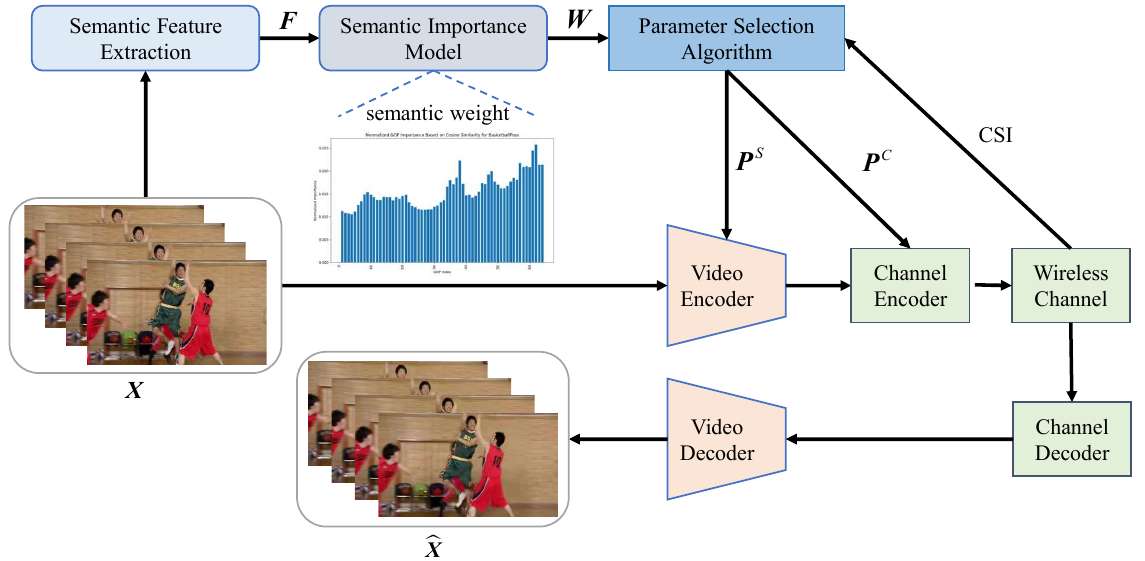}
    \caption{System model of semantic-aware joint source–channel optimization for video communication}
    \label{fig:system model}
\end{figure*}

\subsection{System model}
As shown in Fig. \ref{fig:system model}, we consider an end-to-end (E2E) video semantic communication system, where the video encoder can be either a conventional codec or an AI-based encoder. At the transmitter, a source video is processed in four steps: (\textit{i})~a feature extraction module extracts the semantic features of the video; (\textit{ii})~a semantic importance module evaluates the inter-frame semantic importance of different groups of pictures (GOPs) from these features; (\textit{iii})~based on the semantic importance and the channel state information (CSI), a reinforcement learning algorithm selects the source and channel coding parameters for each GOP; and (\textit{iv})~the video encoder and the channel encoder encode the video with the selected parameters. The encoded bitstream is then transmitted over a wireless channel. At the receiver, a channel decoder recovers the bitstream, and a video decoder matched to the video encoder reconstructs the video.

At the semantic transmitter, the input video frames are $\bm{X} \in \mathbb{R}^{L \times C \times D_{\mathrm{w}} \times D_{\mathrm{h}}}$, where $L$ is the length of the video, $C$ is the number of channels, and $D_{\mathrm{w}}$ and $D_{\mathrm{h}}$ are the width and height of the video frames. The video is divided into $N$ GOPs, where $L = N \times K$, and $\bm{X} = (\bm{X_1}, \bm{X_2}, \dots, \bm{X_{\mathit{N}}})$, where each GOP $\bm{X_i}$ contains $K$ video frames. A pre-trained feature extraction network $f_\omega(\cdot)$ extracts the semantic features $\bm{F} = (\bm{F_1}, \bm{F_2}, \dots, \bm{F_N})$ of the video. Specifically, the $i$-th feature vector $\bm{F_i}$ is obtained as
\begin{equation}
    \bm{F_i} = f_\omega(\bm{X_i}),
    \label{eq:semantic_feature}
\end{equation}
where $\omega$ denotes the parameters of the pre-trained network.
Taking the semantic features $\bm{F}$ as input, a semantic importance model $S(\cdot)$ produces the semantic importance of all GOPs as
\begin{equation}
    \bm{W} = S(\bm{F}),
\end{equation}
where $\bm{W} = (W_1, W_2, \dots, W_N)$ and $W_i$ denotes the semantic importance of GOP $\bm{X_i}$. The model $S(\cdot)$ measures the inter-frame semantic importance of the video and is detailed in Sec.~\ref{sec:SEMANTIC IMPORTANCE EVALUATION}.
The semantic importance and the channel condition jointly determine how each GOP should be coded: A more important GOP or a worse channel calls for stronger protection, whereas a less important GOP or a better channel allows for a lower bit expenditure. Accordingly, each GOP $\bm{X_i}$ is assigned a pair of coding parameters $\bm{P_i} = (P_i^{S}, P_i^{C})$, where the source coding parameter $P_i^{S}$ controls the compression level and the channel coding parameter $P_i^{C}$ controls the degree of error protection. The parameters of all GOPs form the set $\bm{P} = \{\bm{P_i}\}_{i=1}^{N}$, which is determined by a reinforcement learning algorithm $G_\pi(\cdot)$ with policy $\pi$ from the semantic importance $\bm{W}$ and the CSI $\bm{C}$ as
\begin{equation}
    \bm{P} = G_\pi(\bm{W,C}),
    \label{eq:3}
\end{equation}
where $\bm{C} = (C_1, C_2, \dots, C_N)$ and $C_i$ denotes the channel state information (CSI) of the $i$-th timeslot, obtained by transmitting pilot signals at the transmitter and estimating them at the receiver. The design of $G_\pi(\cdot)$ is detailed in Sec.~\ref{sec:PROPOSED ALGORITHM}.

With the selected parameters, the GOPs are encoded one by one. Specifically, the video encoder $E_S$ first encodes GOP $\bm{X_i}$ into a source-coded stream as
\begin{equation}
    \bm{X}_{s,i} = E_S(\bm{X_i}, P_i^{S}),
\end{equation}
where the video encoder can be instantiated as a traditional H.26X codec or a deep neural network-based video encoder. This encoder-agnostic design allows the proposed method to be applied to different video encoders as a plug-in module. The channel encoder $E_C$ then encodes the source-coded stream as
\begin{equation}
    \hat{\bm{X}}_i = E_C(\bm{X}_{s,i}, P_i^{C}),
\end{equation}

The encoded bitstream ${\bm{X}}_i$ is sent over the wireless channel, which is modeled as
\begin{equation}
    \bm{Y_i} = \bm{H} \hat{\bm{X_i}} + \bm{n},
\end{equation}
where $\bm{H}$ represents the wireless channel response, and $\bm{n}$ 
denotes the independent and identically distributed complex Gaussian 
noise following $\mathcal{CN}(0, \sigma^2)$.

To obtain real-time CSI, pilot symbols
$\bm{x}_p$ known to both the transmitter and receiver are transmitted, as
commonly adopted in pilot-assisted channel estimation
\cite{coleri2002channel}. For SNR estimation, the wireless channel $\bm{H}$ is represented by an equivalent complex channel coefficient $h$. The received
pilot signal $\bm{y}_p$ can be expressed as
\begin{equation}
    \bm{y}_p = h\bm{x}_p + \bm{n}.
\end{equation}

Based on the known pilot symbols, the equivalent channel coefficient is
estimated using the least-squares (LS) method as
\begin{equation}
    \hat{h}
    =
    \frac{\bm{x}_p^{H}\bm{y}_p}
    {\bm{x}_p^{H}\bm{x}_p}.
\end{equation}

Based on the estimated channel coefficient, the residual noise is obtained as
\begin{equation}
    \hat{\bm{n}} = \bm{y}_p - \hat{h}\bm{x}_p.
\end{equation}

Then, the SNR is estimated as
\begin{equation}
    \widehat{\mathrm{SNR}}
    =
    \frac{
    \mathbb{E}\left[\left\|\hat{h}\bm{x}_p\right\|^2\right]
    }
    {
    \mathbb{E}\left[\left\|\bm{y}_p-\hat{h}\bm{x}_p\right\|^2\right]
    }.
\end{equation}
CSI estimation is performed once for each GOP. Accordingly, the estimated SNR is used to obtain $C_i$, which represents the CSI associated with the $i$-th timeslot.

Next, the received bitstream is processed by the channel decoder and the video decoder to reconstruct the video, which can be expressed as
\begin{equation}
    \bm{X_i}' = D_S\left(D_C(\bm{Y_i})\right),
\end{equation}
where $D_C(\cdot)$ and $D_S(\cdot)$ denote the channel decoder and the video decoder, respectively. Then we evaluate quality of the reconstructed video. For the $i$-th GOP, the mean square error (MSE) is defined as
\begin{equation}
    \mathrm{MSE}_i = \frac{1}{K D_{\mathrm{h}} D_{\mathrm{w}}} \sum_{k=1}^{K} \sum_{u=1}^{D_{\mathrm{h}}} \sum_{v=1}^{D_{\mathrm{w}}}
    \left(x_{i,k}(u,v) - x_{i,k}'(u,v)\right)^2,
\end{equation}
where $x_{i,k}(m,n)$ and $x_{i,k}'(m,n)$ denote the pixel values at location $(m,n)$ in the $k$-th frame of $\bm{X}_i$ and $\bm{X}_i'$, respectively. 
Accordingly, the Peak Signal-to-Noise Ratio (PSNR) of the $i$-th GOP is given by
\begin{equation}
    Q_i = 10 \log_{10} \left( \frac{V_{\max}^2}{\mathrm{MSE}_i} \right),
\end{equation}
where $V_{\max}$ denotes the maximum possible pixel value.

Different GOPs contribute unequally to the overall semantic perception of the video, whereas conventional distortion metrics such as PSNR and SSIM treat all GOPs equally and therefore cannot adequately characterize their semantic significance. Inspired by \cite{erfurt2019study, gao2025cross, wang2023semantic}, we therefore adopt the semantic importance weighted peak signal-to-noise ratio (WPSNR) as the semantic-aware video reconstruction quality metric. Specifically, we assign a semantic weight to each GOP and define the weighted quality metric as
\begin{equation}
     Q_{i}^{w} = W_i \cdot Q_i,
     \label{eq:wpsnr}
\end{equation}
where $W_i$ denotes the semantic weight of the $i$-th GOP, $Q_i$ denotes the PSNR of the $i$-th GOP, and $Q_i^{w}$ denotes the corresponding weighted PSNR. 
Accordingly, WPSNR provides a semantic-aware measure of reconstructed video quality by emphasizing GOPs with higher semantic importance.



\subsection{Problem formulation}

Based on the above semantic-aware video quality metric, we formulate a joint source-channel optimization problem to maximize the overall weighted reconstruction quality under a bitrate constraint. The optimization problem is expressed as
\begin{equation}
    \begin{aligned}
        \max_{\bm{P}} \quad & \sum_{i=1}^{N} Q_i^{w} \\
        \text{s.t.} \quad
        & R(\bm{P}) \leq R_{\max}, \\
        & P_{i}^{S} \in \mathcal{P}^{S}, \quad \forall i=1,\ldots,N, \\
        & P_{i}^{C} \in \mathcal{P}^{C}, \quad \forall i=1,\ldots,N,
    \end{aligned}
    \label{eq:optimization_problem}
\end{equation}
where $R(\bm{P})$ denotes the total bitrate corresponding to $\bm{P}$, $R_{\max}$ is the maximum allowable bitrate, and $\mathcal{P}^{S}$ and $\mathcal{P}^{C}$ represent the feasible sets of the source encoding parameter and channel encoding parameter, respectively.

The formulated optimization problem is difficult to solve directly due to its non-convex nature. 
Specifically, the joint selection of source and channel coding parameters involves a complex rate-distortion tradeoff that depends on both video semantic importance and time-varying channel conditions. 
Moreover, the mapping from coding parameters to reconstructed video quality is highly nonlinear and cannot be explicitly characterized by a closed-form mathematical expression. 
To address this issue, we adopt a heuristic learning-based approach and employ deep reinforcement learning (DRL) to learn an effective joint source-channel optimization policy under varying video semantics and channel states.

\section{Inter-frame Semantic Importance Evaluation}\label{sec:SEMANTIC IMPORTANCE EVALUATION}

\begin{figure*}[t]
    \centering
    \includegraphics[width=0.8\linewidth, height=0.26\textheight, keepaspectratio]{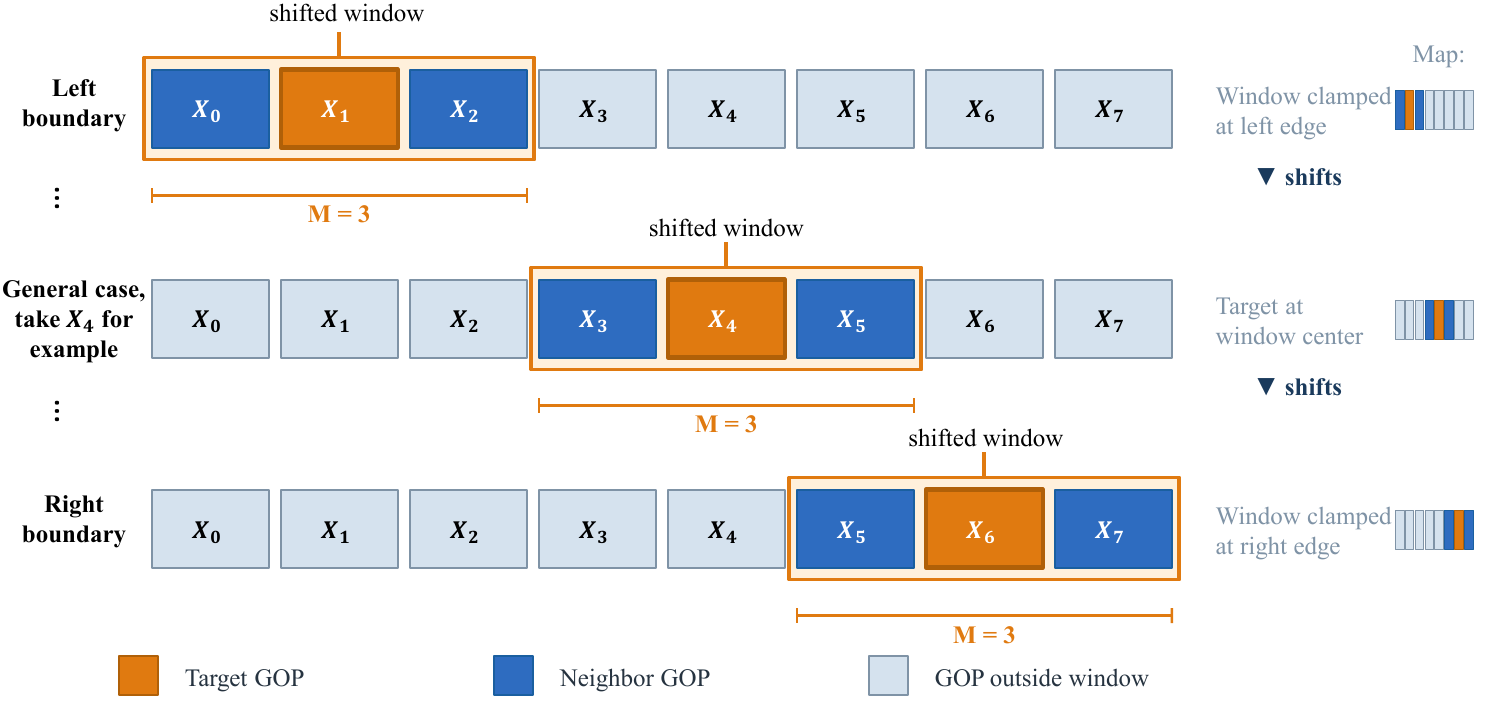}
    \caption{Illustration of the proposed inter-GOP semantic importance calculation method.}
    \label{fig:semantic_importance}
\end{figure*}

The temporal semantic importance of each GOP is evaluated via an inter-frame 
approach. We first extract semantic features from video GOPs 
and subsequently model the cosine similarity at feature level between adjacent GOPs.

\subsection{Shifted-Window Feature Construction}
The extracted features effectively preserve both the spatial structure and temporal dynamics of the video, thereby providing rich semantic information for subsequent modeling.
To capture the inter-frame semantic relevance among adjacent GOPs, we adopt a shifted-window strategy, inspired by the shifted window mechanism in Swin Transformer \cite{liu2021swin}, to perform local dependency analysis in the temporal domain. As illustrated in Fig.~\ref{fig:semantic_importance}, the proposed method calculates the semantic importance of each GOP by measuring its semantic similarity with neighboring GOPs within the local window.
Specifically, for each target GOP $\bm{X}_i$, the local feature window is constructed to include the semantic features of $M$ consecutive GOPs centered around the target GOP as much as possible.
The corresponding index set is denoted by $\mathcal{M}(i)$ and defined as
\begin{equation}
\mathcal{M}(i)=\left\{j \in \mathbb{Z} \,\middle|\, b_{\mathrm{L}}^{(i)} \leq j \leq b_{\mathrm{R}}^{(i)} \right\},
\end{equation}
where $b_{\mathrm{L}}^{(i)}$ and $b_r$ denote the left and right boundaries of the window, respectively, subject to the constraints
\begin{equation}
0 \leq b_{\mathrm{L}}^{(i)} \leq i \leq b_{\mathrm{R}}^{(i)} \leq N-1,\quad b_{\mathrm{R}}^{(i)}-b_{\mathrm{L}}^{(i)}+1=M.
\end{equation}

In this way, the target GOP is positioned as close to the center of the window as possible, while the window size $M$ remains constant. By operating within local shifted windows, this strategy captures the temporal semantic dependencies among adjacent GOPs, thereby circumventing redundant global computations.

\subsection{Cosine-Similarity-Based Importance Metric}

To quantify the temporal semantic importance of video content, we propose 
a cosine similarity-based inter-frame semantic importance measurement method. 
For two GOPs semantic features $F_i$ and $F_j$, the cosine similarity is defined as
\begin{equation}
    \mathrm{sim}(F_i,F_j)=\frac{\left\langle F_i,F_j \right\rangle}{\left\|F_i\right\|\left\|F_j\right\|},
\end{equation}
where $\langle \cdot, \cdot \rangle$ denotes the inner product. Since the semantic features extracted by the neural network are non-negative, and zero feature vectors do not arise in practice for natural video content, the cosine similarity is guaranteed to be strictly positive, i.e., $\mathrm{sim}(F_i,F_j)\in(0,1]$. Based on cosine similarity, the semantic importance of the $i$-th GOP is defined as the inverse of the average cosine similarity between the target GOP and the other GOPs within the window,
\begin{equation}
\label{eq:semantic_importance}
W_i=\frac{1}{\frac{1}{M-1}\sum_{\substack{j\in\mathcal{M}(i)\\j\neq i}}\mathrm{sim}(F_i,F_j)}.
\end{equation}

This definition implies that a GOP with lower semantic similarity to its neighboring GOPs is assigned a higher semantic importance, since it contains more distinct semantic information in the temporal sequence.

By sliding the local window over the entire video, the inter-GOP semantic importance vector of the video sequence can be obtained as
\begin{equation}
\bm{W}=\left(W_1,W_2,\ldots,W_N\right).
\end{equation}

The resulting semantic importance vector is then used to guide subsequent semantic-aware source-channel parameter optimization. In our previous work~\cite{zhu2026semantic}, semantic importance was used to adaptively adjust the keyframe interval.

\section{MPPO Algorithm for Joint Source–Channel Optimization} \label{sec:PROPOSED ALGORITHM}
In this section, we propose a novel DRL algorithm called multi-actor proximal policy optimization (MPPO), which incorporates multiple actor networks and action spaces, to address the joint source-channel optimization problem based on inter-frame semantic importance.

\subsection{DRL-based Parameter Selection Algorithm}
In this subsection, we begin by discussing the key components of the MPPO algorithm and the step-by-step process of
utilizing it to optimize the video joint source-channel optimization strategy. Subsequently, we delve into the complexity analysis of the proposed MPPO algorithm.
We first present the main components of the proposed MPPO algorithm and detail the procedure for employing it to optimize the joint source-channel optimization strategy for video transmission. We then provide a computational complexity analysis of the proposed MPPO algorithm.

\textit{1) Components of the MPPO Algorithm:}
In this part, we provide a comprehensive description of the key components of the proposed MPPO algorithm. Once the video semantic features have been extracted and the corresponding inter-frame semantic importance has been computed, the joint source-channel optimization process is formulated as a Markov decision process (MDP). In this process, each GOP is sequentially encoded from $i=1$ to $i=N$ according to the selected source and channel coding parameters.
The MPPO algorithm consists of four essential components: a) action space, b) state space, c) reward function, and d) DRL agent. These components are outlined as follows:

\textit{a) Action space:} The agent action consists of two parts, namely the source coding action and the channel coding action. Since source coding and channel coding focus on coding efficiency and transmission reliability, respectively, their optimization objectives are inherently different. Therefore, the proposed MPPO framework employs two actor networks to learn the corresponding policies.
The source coding action is selected from the action space $\mathcal{A}_{S}=\{CR_{\min}, \ldots, CR_{\max}\}$, where $CR$ denotes the compression rate. Similarly, the channel coding action is selected from the action space $\mathcal{A}_{C}=\{r_{\min}, \ldots, r_{\max}\}$, where $r$ denotes the channel coding rate, and $r_{\min}$ and $r_{\max}$ represent the minimum and maximum channel coding rates, respectively.
Here, $a_{S}^{(i)}\in\mathcal{A}_{S}$ and $a_{C}^{(i)}\in\mathcal{A}_{C}$ denote the specific source coding action and channel coding action selected at the $i$-th decision step, respectively. At each step, the two actor networks output the source and channel coding actions, after which source coding is performed first, followed by channel coding.

\textit{b) State space:} The state of the $i$-th GOP is defined as $s^{(i)} = [W_i, \bar{W}_i, f_i^{W}, \bar{\gamma}_i, f_i^{\gamma}]$, where $W_i$ denotes the semantic weight of the current GOP, $\bar{W}_i$ denotes its normalized value in the range $[0,1]$, $f_i^{W} \in \{-1,1\}$ is a binary semantic-weight flag, which is determined by whether the semantic weight of the $i$-th GOP is larger than the median value of all semantic weights, $\bar{\gamma}_i$ denotes the normalized channel state derived from the SNR, and $f_i^{\gamma} \in \{-1,1\}$ is a binary channel-state flag dynamically determined according to the SNR, where the threshold is set to the integer SNR point at which the decoding-failure probability is approximately $1/2$. In this way, all state variables are scaled to a comparable range, which facilitates the observation and learning process of the agent.

\textit{c) Reward function:} The reward function is defined as
\begin{equation}
\label{eq:reward_function}
R_i = c_1 \cdot Q_i^{w} - c_2 \cdot d_i + c_3 \cdot R_i^{\mathrm{snr}} + c_4 \cdot R_i^{\mathrm{weight}} - \rho \cdot \delta_i,
\end{equation}
where $d_i$ denotes the amount of transmitted data, $R_i^{\mathrm{snr}}$ and $R_i^{\mathrm{weight}}$ denote the reward terms associated with the SNR flag and the semantic-weight flag, respectively. 
This reward function encourages the agent to maximize the reconstruction quality while controlling the transmission cost.
Specifically, $R_i^{\mathrm{weight}}$ is defined as $R_i^{\mathrm{weight}}=f_i^{W}\left(a_{S}^{(i)}-\tilde{a}_{S}\right)$, and $R_i^{\mathrm{snr}}$ is defined as $R_i^{\mathrm{snr}}=f_i^{\gamma}\left(a_{C}^{(i)}-\tilde{a}_{C}\right)$, where $\tilde{a}_{S}$ and $\tilde{a}_{C}$ denote the median values of the source coding action set $\mathcal{A}_{S}$ and the channel coding action set $\mathcal{A}_{C}$, respectively.
Moreover, $\delta_i = 1$ if decoding failure occurs at step $i$ and $\delta_i = 0$ otherwise, while $\rho > 0$ is a predefined penalty constant. The term $-c_2d_i$ serves as a penalty on the amount of transmitted data. Since this penalty is accumulated over each training episode, selecting actions that incur higher transmission costs without yielding commensurate improvements in reconstruction quality results in a lower cumulative reward. Consequently, the agent learns to avoid such inefficient actions and select source--channel coding parameters that maintain the total bitrate within the constraint $R(\boldsymbol{P}) \leq R_{\max}$. 
The coefficients $c_1, c_2, c_3, c_4$ are hyperparameters that balance the 
respective reward terms and are found insensitive to the choice of dataset 
in practice. By aligning the step-wise reward with the objective in (\ref{eq:optimization_problem}), the proposed MPPO algorithm solves the formulated optimization problem through cumulative reward maximization.

\textit{d) DRL Agent:} The agent is the decision-making entity that interacts with the environment, and is implemented in practice by a set of neural networks. Specifically, it  maintains two separate policies, 
$\pi_S(a_S^{(i)} \mid s^{(i)})$ and $\pi_C(a_C^{(i)} \mid s^{(i)})$, 
governing the selection of source coding and channel coding actions, respectively. 
At each step, the agent observes the current state $s^{(i)}$, samples actions 
from the two policies, and receives the reward $R^{(i)}$ from the environment. 
By iteratively updating both policies through reward feedback and state transitions, 
the agent gradually learns the optimal source encoding and channel coding strategy. In the proposed MPPO algorithm, the agent is implemented with two actor networks and one shared critic network, corresponding to the two decoupled action spaces.

\textit{2) Network Architecture:}

\begin{figure*}[t]
    \centering
    \includegraphics[width=0.8\textwidth, height=0.26 \textheight, keepaspectratio]{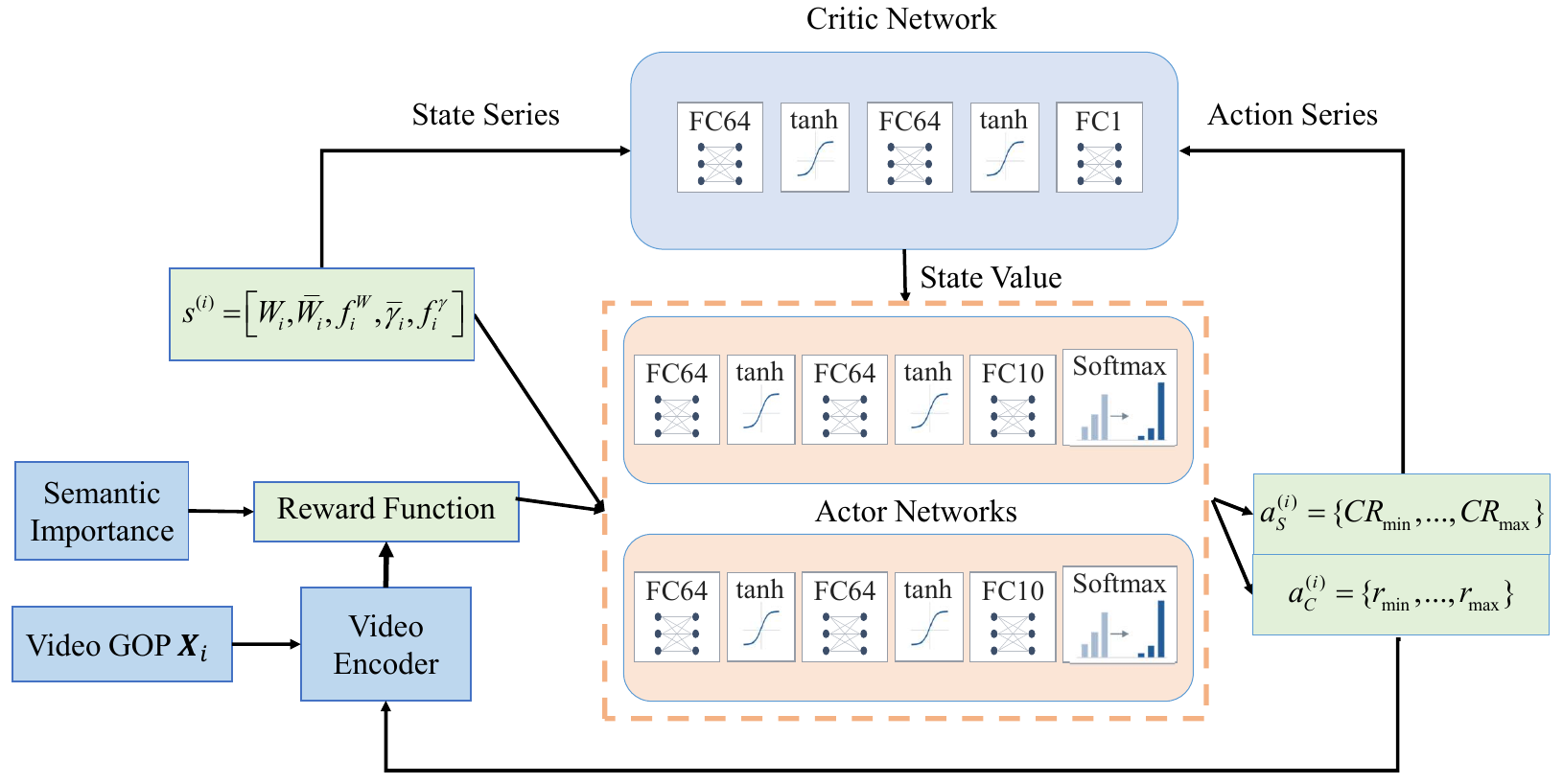}
    \caption{Architecture of semantic-aware joint source–channel optimization for video communication}
    \label{fig:architecture_MPPO}
\end{figure*}

As illustrated in Fig.~\ref{fig:architecture_MPPO}, the architecture of the proposed MPPO mainly consists of two actor networks and one critic network. The two actor networks have the same structure, but are used to generate different actions, namely the source coding action $a_S^{(i)}$ and the channel coding action $a_C^{(i)}$, respectively. Each actor network takes the input state $s^{(i)}$ and consists of three consecutive fully connected (FC) layers. The first two FC layers are followed by the Tanh activation function, while the final layer employs a Softmax function to generate the predicted probability distribution over the candidate actions. Based on the output distributions of the two actor networks, the source coding action and channel coding action are selected for the $i$-th GOP. 

The critic network is used to evaluate the current decision and also consists of three FC layers. The first two layers are followed by the Tanh activation function, while the final layer outputs the state value, which serves as an estimate of the quality of the selected actions at the current state. In this way, the two actor networks and the critic network work cooperatively to determine the action selection process and evaluate the corresponding decision quality, thereby optimizing the joint source-channel optimization policy.

\textit{3) Training Procedure:}
We next introduce the training procedure of the proposed MPPO algorithm. First, the current state $s^{(i)}$ is fed into the MPPO agent, which learns the source coding policy $\pi_S$ and the channel coding policy $\pi_C$ and accordingly outputs actions $a_S^{(i)}$ and $a_C^{(i)}$ for the source encoder and channel encoder, respectively. Based on the selected coding parameters, the video is successively processed by source coding and channel coding to generate the encoded bitstream, which is then transmitted over the wireless channel. After transmission, channel decoding and source decoding are performed sequentially to reconstruct the video. Then, the WPSNR of the reconstructed video is calculated to evaluate the video quality. Based on the transmission result and reconstruction quality, the reward is obtained and used to guide the update of the policy networks.

During the parameter updating phase, the proposed MPPO framework updates one critic network with parameters $\psi$ and two actor networks with parameters $\theta$, following the multi-actor reinforcement learning architecture in \cite{yang2026safety}. For each actor network, the parameter updating procedure is performed as follows. We first compute the discounted cumulative reward at timestep $t$, denoted as $R^t$, which can be expressed as
\begin{equation}
\label{eq:cumulative_reward}
R^t=\sum_{k=1}^{t}\eta^{t-k}r^k,
\end{equation}
where $\eta$ denotes the discount factor and $r^k$ is the immediate reward at timestep $k$. Then, the estimator of the advantage function at timestep $t$, denoted as $\hat{\phi}^t$, is calculated by
\begin{equation}
\label{eq:advantage_function}
\hat{\phi}^t = R^t - e^t,
\end{equation}
where $e^t$ is the state value estimated by the critic network. The estimator $\hat{\phi}^t$ reflects the advantage of the selected action over the expected return at the current state.

Next, the probability ratio between the new policy and the old policy is computed as
\begin{equation}
p^t(\theta)=\frac{\pi_{\theta}(a^t|s^t)}{\pi_{\theta_{\mathrm{old}}}(a^t|s^t)},
\end{equation}
where $\theta_{\mathrm{old}}$ denotes the vector of policy parameters before the update. In line with \cite{schulman2017proximal}, the main objective of the actor network is
\begin{equation}
\label{eq:ppo_surrogate}
L_1(\theta)=\mathbb{E}\left[\min\left(p^t(\theta)\hat{\phi}^t,\mathrm{clip}\left(p^t(\theta),1-\epsilon,1+\epsilon\right)\hat{\phi}^t\right)\right],
\end{equation}
where $\epsilon$ is clip parameter and $\mathbb{E}$ is the expectation operator. The clipping function $\mathrm{clip}(\cdot)$ is used to constrain the probability ratio within a predefined range, so as to avoid excessively large policy updates and improve the stability of training.

Following the work in \cite{schulman2017proximal}, the overall objective further incorporates a value function loss term and an entropy bonus term, which can be written as
\begin{equation}
\label{eq:ppo_loss}
L(\theta)=\hat{\mathbb{E}}_t\left[L_1(\theta)-c_5L_t^{VF}(\theta)+c_6S[\pi_\theta](s^t)\right],
\end{equation}
where $c_5$ and $c_6$ are coefficients, $S$ denotes the entropy bonus, and $L_t^{VF}(\theta)$ is the value function loss. In our implementation, the value function loss is defined as $L_t^{VF}(\theta)=\left(e^t-R^t\right)^2$. The second term improves the accuracy of state-value estimation, while the third term encourages policy exploration.

Finally, the policy parameters are updated by gradient descent on the sampled trajectories according to
\begin{equation}
\label{eq:policy_update}
\theta^{(k)} \leftarrow \theta^{(k-1)}-\delta \nabla_{\theta} O(\theta),
\end{equation}
where $\theta^{(k)}$ denotes the policy parameters at the $k$-th iteration and $\delta$ is the learning rate.

By iteratively updating the policy until convergence, the proposed MPPO algorithm can learn the mapping from source encoding and channel coding parameters to reconstructed video quality. In this way, it is able to determine an effective policy for achieving the optimal rate-distortion tradeoff. The detailed training procedure of the proposed MPPO algorithm is summarized in Algorithm~\ref{alg:mppo}.

\begin{algorithm}[t]
\caption{Training Process of the Proposed MPPO Algorithm}
\label{alg:mppo}
\begin{algorithmic}[1]
\STATE \textbf{Input:} Video dataset $\mathcal{D}$, training epoch $E_p$, learning rate $\delta$, discount factor $\eta$, hyper-parameters $\epsilon$, $c_5$, and $c_6$, pre-trained semantic feature extractor, source encoder, and channel encoder.
\STATE \textbf{Output:} Policy network parameters $\theta$ and critic network parameters $\psi$ 
\REPEAT
    \STATE Sample a batch of video sequences from dataset $\mathcal{D}$.
    \STATE Extract semantic features and compute GOP semantic importance by (\ref{eq:semantic_importance}).
    \STATE Collect trajectories using the old policies.
    \STATE Compute the cumulative reward by (\ref{eq:cumulative_reward}) and the loss function by (\ref{eq:ppo_loss}).
    \STATE Update the network parameters by (\ref{eq:policy_update}).
\UNTIL{the loss function in (\ref{eq:ppo_loss}) converges.}
\end{algorithmic}
\end{algorithm}

\subsection{Complexity Analysis}
The computational complexity of the proposed MPPO framework arises from three components, namely semantic feature extraction, inter-frame semantic importance evaluation, and the MPPO algorithm.

For the first part, semantic features are extracted by pretrained deep neural network, the complexity of feature extraction for one GOP can be approximated as $\mathcal{O}(D_{\mathrm{h}} D_{\mathrm{w}})$. Therefore, for a video sequence containing $N$ GOPs, the overall complexity of semantic feature extraction is
\begin{equation}
\mathcal{O}(N D_{\mathrm{h}} D_{\mathrm{w}}).
\end{equation}

For the second part, inter-frame semantic importance is computed based on cosine similarity. Let $d_f$ denote the dimension of the semantic feature and $M$ denote the semantic window size. Since each GOP needs to compute cosine similarity with the other $M-1$ GOPs in the window, the complexity for one GOP is $\mathcal{O}((M-1)d_f)$. Hence, for all $N$ GOPs, the complexity of semantic importance evaluation is
\begin{equation}
\mathcal{O}(N(M-1)d_f).
\end{equation}

For the third part, let $d_s$ and $d_a$ denote the dimensions of the state space and action space, respectively. Let $H_l$ denote the number of neurons in the $l$-th layer of the neural network, and let $L$ denote the number of layers. The computational complexity of the MPPO-based decision process for one GOP can be approximated as~\cite{liu2024ofdm}
\begin{equation}
\mathcal{O}\left(d_s d_a \prod_{l=1}^{L} H_l\right).
\end{equation}
Accordingly, for a video sequence with $N$ GOPs, the complexity of this part is
\begin{equation}
\mathcal{O}\left(N d_s d_a \prod_{l=1}^{L} H_l\right).
\end{equation}

Therefore, the total computational complexity of the proposed framework can be expressed as
\begin{equation}
\mathcal{O}\left(ND_{\mathrm{h}} D_{\mathrm{w}} + N(M-1)d_f + N d_s d_a \prod_{l=1}^{L} H_l\right).
\end{equation}

\section{Simulation Results and Analysis}\label{sec:SIMULATION RESULTS AND ANALYSIS}
In this section, we perform a comprehensive series of simulations aimed at validating the efficacy of both the proposed SAJSCO scheme and the MPPO algorithm.

\subsection{Simulation Setup}
\textit{1) Video Dataset:} We use the ActivityNet dataset as the training dataset for the proposed MPPO algorithm. ActivityNet is a large-scale benchmark for human activity understanding, consisting of untrimmed videos collected from YouTube and covering 203 activity categories. Since the videos are collected from real-world web sources, their spatial resolutions are not fixed and vary across samples~\cite{caba2015activitynet}. In our experiments, the training and test subsets of ActivityNet are divided with a ratio of 5:1. For testing, we also use the HEVC test dataset, where the selected sequences belong to Class D with a resolution of $416\times240$ according to the HEVC common test conditions. 

\textit{2) Channel Environment:} 
The channel dataset used in this work is RadioML2016.10a~\cite{o2016radio}, which is a widely used benchmark dataset for wireless signal modulation analysis under different channel conditions. Specifically, we select one SNR sequence from the RadioML2016.10a dataset and use its first 64 SNR values to model the time-varying channel condition. The selected SNR values are then normalized and remapped to the range of 0--20 dB according to the considered simulation setting. In addition, LDPC coding is employed for channel encoding and decoding to enhance transmission reliability over the wireless channel.

\textit{3) Semantic Extraction Module:} 
The video semantic extractor is built upon the MobileNetV2 architecture~\cite{sandler2018mobilenetv2}, which employs inverted residual blocks, linear bottlenecks, and depthwise separable convolutions to reduce computational complexity. It efficiently extracts discriminative frame-level semantic features.

\textit{4) Baselines:} 
The baselines are categorized as follows. (\textit{i})~Non-semantic baselines: H.265 and DCVC-RT~\cite{jia2025towards} with fixed coding rate configurations are included as representatives of non-semantic video coding. Both encoders exploit intra-frame and inter-frame redundancy for compression, yet their fixed-parameter configurations allocate resources uniformly without considering the semantic importance of video content. The proposed SAJSCO employs a single MPPO policy, which is trained once and subsequently integrated with both H.265 and DCVC-RT as a plug-in module without encoder-specific retraining or fine-tuning, thereby enabling temporal semantic-aware adaptive parameter selection across heterogeneous video encoders. (\textit{ii})~Spatial-semantic baseline: SwinJSCC~\cite{yang2024swinjscc}, a representative analog joint source-channel coding (JSCC) method, is included for comparison. SwinJSCC considers only spatial semantics at the intra-frame level and does not capture temporal semantic dependencies across frames. For a fair comparison, the first frame of each GOP is selected as the key frame and fed into SwinJSCC as an individual image, with the channel bandwidth ratio kept consistent with that of the proposed method.

In the simulations, the source coding action space is instantiated by five CR levels, i.e., 
$a_{S}^{(i)} \in \{35, 37, 40, 43, 45\}$, 
which correspond to different source compression rates. 
The channel coding action space is instantiated by three LDPC coding rates, i.e., 
$a_{C}^{(i)} \in \{1/3, 1/2, 2/3\}$. The modulation scheme employed is 16QAM. The quality of the reconstructed videos is evaluated by using PSNR, WPSNR, LPIPS~\cite{zhang2018unreasonable}, and semantic importance weighted LPIPS(WLPIPS). Similar to the computation of WPSNR in \eqref{eq:wpsnr}, WLPIPS is obtained by weighting the LPIPS of each GOP according to its semantic importance. PSNR and WPSNR reflects the distortion at the pixel level, whereas LPIPS and WLPIPS measures perceptual similarity from the perspective of deep feature representations. The simulations and experiments are performed by the computer with Ubuntu20.04 + CUDA12.9, and the selected deep learning framework is Pytorch. The hardware configuration includes a single NVIDIA Tesla V100 GPU and an Intel(R) Xeon(R) Gold 6240 CPU @ 2.60GHz. Other simulation parameters are summarized in Table~\ref{tab:simulation_parameters}.

\begin{table}[t]
\caption{Simulation Parameters}
\label{tab:simulation_parameters}
\centering
\resizebox{\columnwidth}{!}{%
\begin{tabular}{cccc}
\toprule
\textbf{Parameter} & \textbf{Value} & \textbf{Parameter} & \textbf{Value} \\
\midrule
Number of GOPs, $N$              & 64               & Shifted-window size, $M$      & 8  \\
Training episodes                & 500              & PPO update epochs, $E_p$      & 8  \\
Discount factor, $\eta$          & 0.999            & Clip parameter, $\epsilon$    & 0.2 \\
Optimizer                        & Adam             & Penalty constant, $\rho$ & 50 \\
Actor learning rate, $\delta_a$  & $1\times10^{-3}$ & Critic learning rate, $\delta_c$ & $3\times10^{-3}$ \\
Hyper-parameter, $c_1$           & 10               & Hyper-parameter, $c_2$        & 0.3 \\
Hyper-parameter, $c_3$           & 200              & Hyper-parameter, $c_4$        & 10  \\
Hyper-parameter, $c_5$           & 0.5              & Hyper-parameter, $c_6$        & 0.5 \\
\bottomrule
\end{tabular}}
\end{table}

\subsection{Training Convergence and Ablation Study}

\begin{figure}[t]
    \centering
    \includegraphics[width=\linewidth]{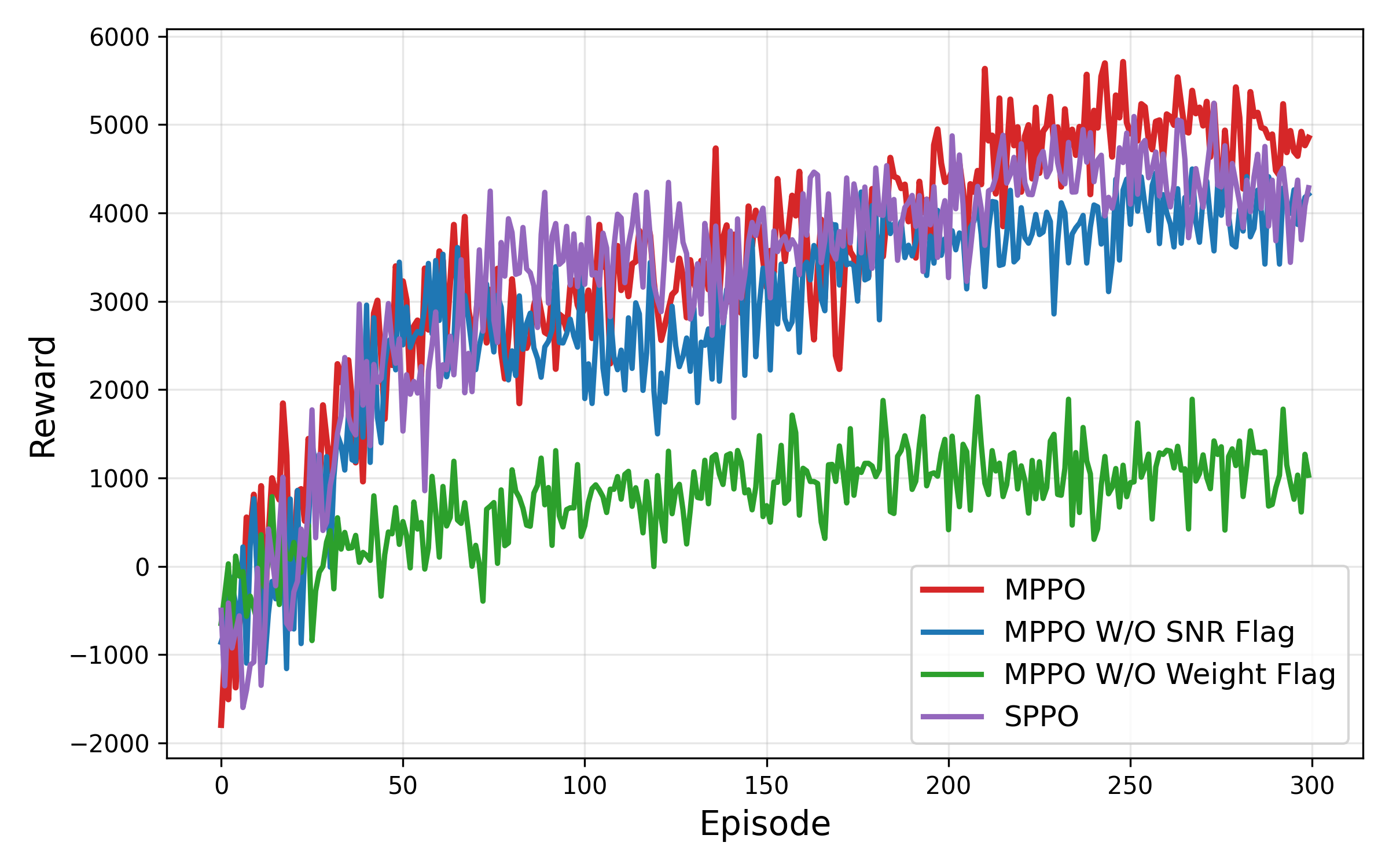}
    \caption{Ablation study of the proposed MPPO algorithm.}
    \label{fig:reward_ablation}
    \vspace{-0.5em}
\end{figure}

To evaluate the effectiveness of both the state space design and the multi-actor policy architecture, we conduct an ablation study during the training stage. Specifically, two state-space variants are constructed by separately removing the semantic-weight flag $f_i^{W}$ and the channel-state flag $f_i^{\gamma}$. In addition, the proposed MPPO is compared with a single-actor PPO (SPPO), in which a single actor jointly selects the source and channel coding actions. All variants are evaluated under the same training settings to ensure a fair comparison. The convergence process over the first 300 training episodes is illustrated in Fig.~\ref{fig:reward_ablation}.
It can be observed that the proposed MPPO converges faster and achieves a higher and more stable reward than all ablated variants. Removing either $f_i^{W}$ or $f_i^{\gamma}$ degrades the training performance. Moreover, MPPO consistently outperforms SPPO, which can be attributed to the distinct functional roles of source and channel coding: Source coding governs compression efficiency, whereas channel coding is primarily responsible for ensuring transmission reliability. Unlike a single actor that may struggle to disentangle the heterogeneous roles of the two actions, the multi-actor architecture allows each actor to specialize in its respective decision while optimizing a shared reward.

\subsection{Video Transmission Under Fixed CSI}

\begin{figure*}[t]
    \centering
    \subfloat[PSNR versus SNR]{
        \includegraphics[width=0.48\linewidth]{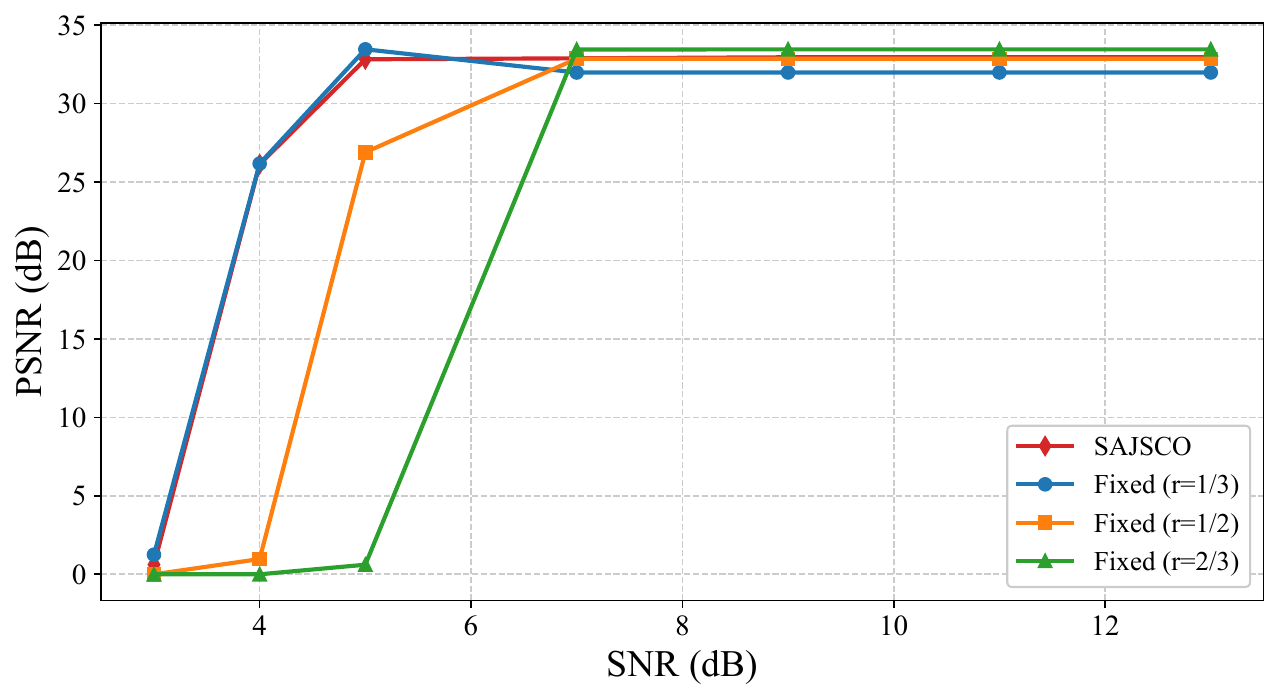}
        \label{fig:act_h265_snr_psnr}
    }
    \hfill
    \subfloat[LPIPS versus SNR]{
        \includegraphics[width=0.48\linewidth]{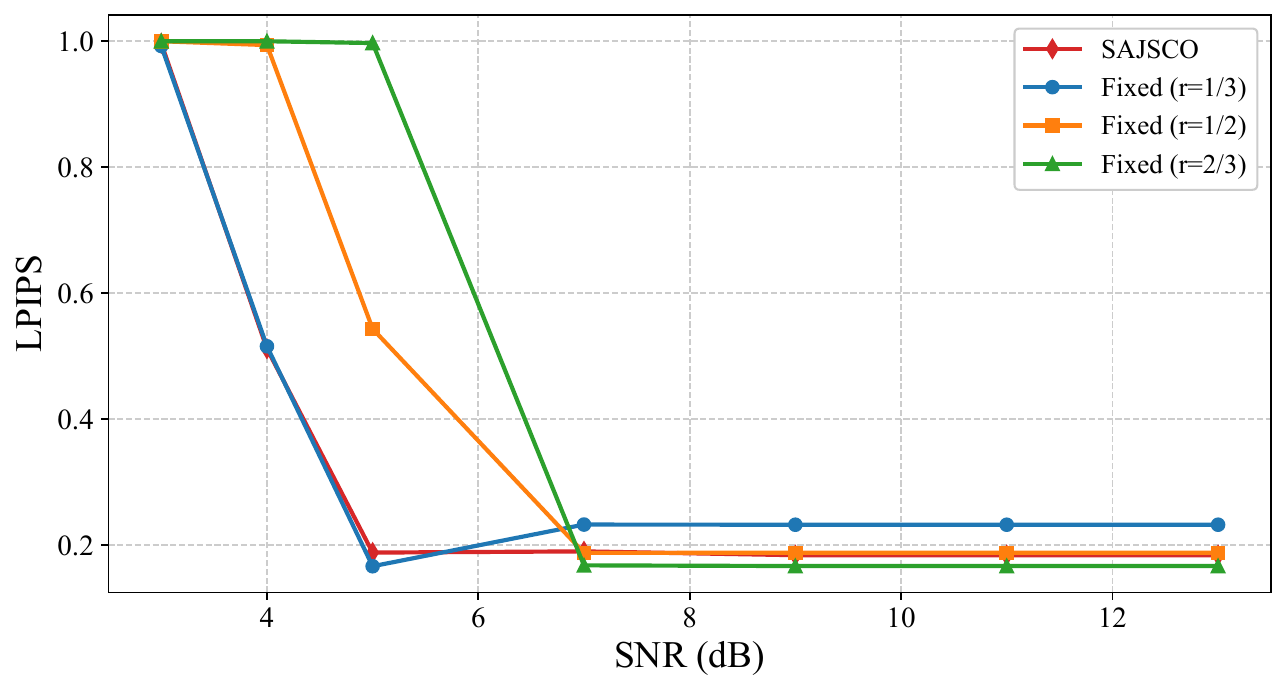}
        \label{fig:act_h265_snr_lpips}
    }
    \caption{Performance comparison under different SNR conditions on the ActivityNet dataset with the H.265 encoder.}
    \label{fig:act_h265_snr}
\end{figure*}

We employ the H.265 video encoder on Activitynet dataset and compare against methods with fixed source-channel coding parameters. Specifically, three baselines are constructed using the same three LDPC code rates as the action space, and the CR is adjusted to match the bitrate of the proposed method. The PSNR and LPIPS results versus SNR are shown in Fig.~\ref{fig:act_h265_snr}. The reconstructed video quality exhibits an evident cliff effect as the SNR varies. This is mainly because the conventional H.265 bitstream is highly sensitive to channel errors. Once critical bits are corrupted during transmission, the decoder may fail, resulting in a sharp performance degradation. The proposed SAJSCO method consistently achieves performance close to the best fixed coding strategy over a wide SNR range, demonstrating strong adaptability to varying channel conditions. More specifically, under low-SNR conditions, the proposed method tends to allocate more bandwidth to improve transmission robustness. As the SNR increases, it gradually reduces the bandwidth while maintaining competitive reconstruction quality, therefore a slight performance decline can also be observed in the high-SNR region.

\subsection{Video Transmission Under Time-Varying CSI} 
In this experiment, the previously introduced RadioML channel dataset is adopted to more realistically simulate the time-varying wireless channel during video transmission. Specifically, one SNR value is assigned to each timestep, and after $N$ transmission steps, the average PSNR, WPSNR, LPIPS, and WLPIPS of the entire video transmission process are calculated. For each encoder, three fixed channel coding rates are selected, and the rate set is identical to the candidate channel coding rates in the action space of the proposed MPPO algorithm, so as to ensure a fair comparison. 

\begin{figure}[t]
    \centering
    \subfloat[PSNR versus bitrate]{
        \includegraphics[width=0.92\linewidth]{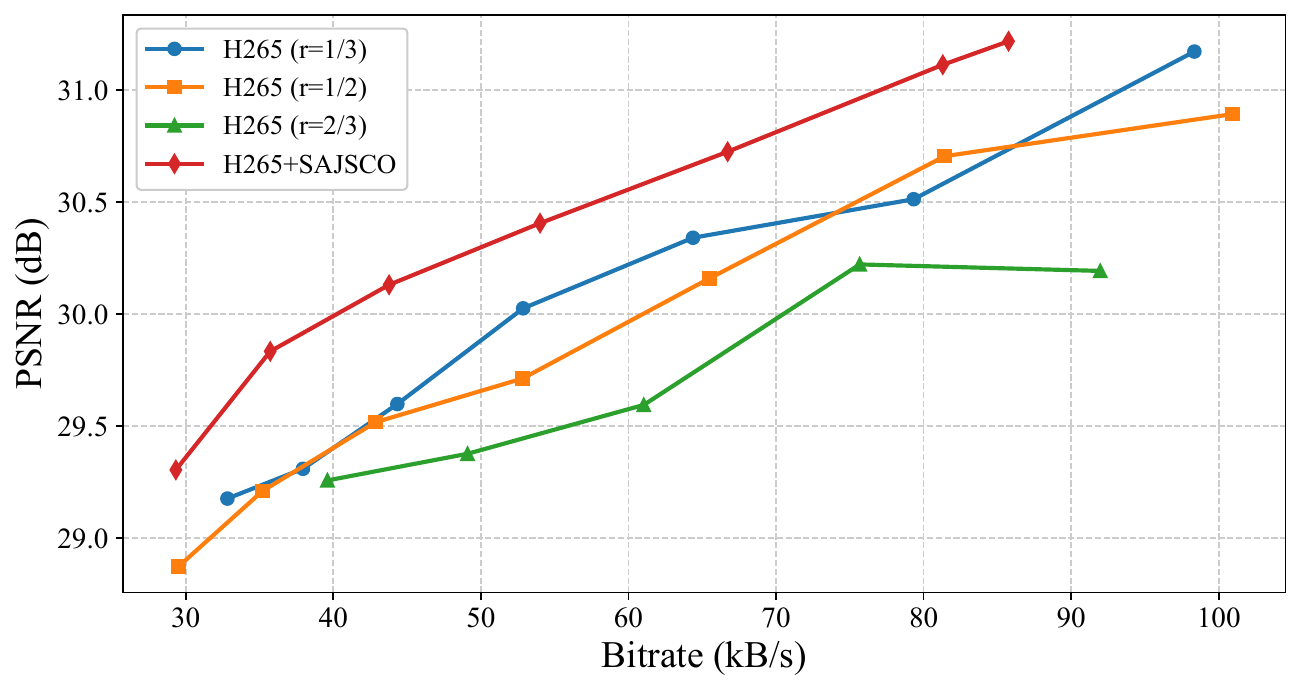}
        \label{fig:act_h265_psnr}
    }\\
    \subfloat[LPIPS versus bitrate]{
        \includegraphics[width=0.95\linewidth]{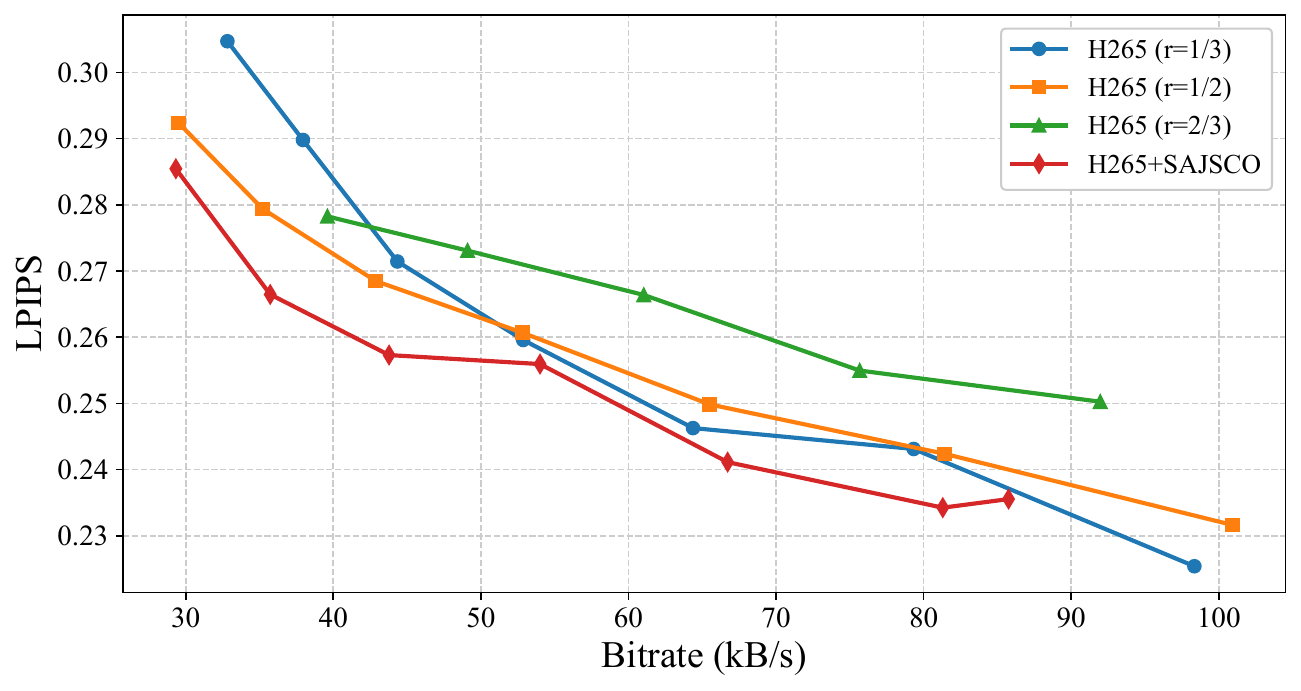}
        \label{fig:act_h265_lpips}
    }
    \caption{PSNR and LPIPS performance versus bitrate on the ActivityNet dataset with the H.265 encoder under the RadioML channel setting.}
    \label{fig:act_h265_psnr_lpips}
\end{figure}

Fig.~\ref{fig:act_h265_psnr_lpips} shows the PSNR and LPIPS results versus bitrate on the ActivityNet dataset, where H.265 is adopted as the video encoder and RadioML is used to model the wireless channel. For PSNR, SAJSCO achieves BD-rate reductions of $22.74\%$, $28.92\%$, and $52.87\%$ compared with the three non-semantic baselines, respectively. For LPIPS, the proposed method reduces the BD-rate by $17.69\%$, $15.73\%$, and $36.09\%$ compared with the three non-semantic baselines, respectively. Moreover, SAJSCO also improves the average PSNR over the overlapping bitrate range while reducing LPIPS, indicating better perceptual quality. These results confirm that the proposed method can effectively adapt the source and channel coding parameters according to the channel condition.

\begin{figure*}[t]
    \centering
    \subfloat[PSNR versus bitrate]{
        \includegraphics[width=0.48\linewidth]{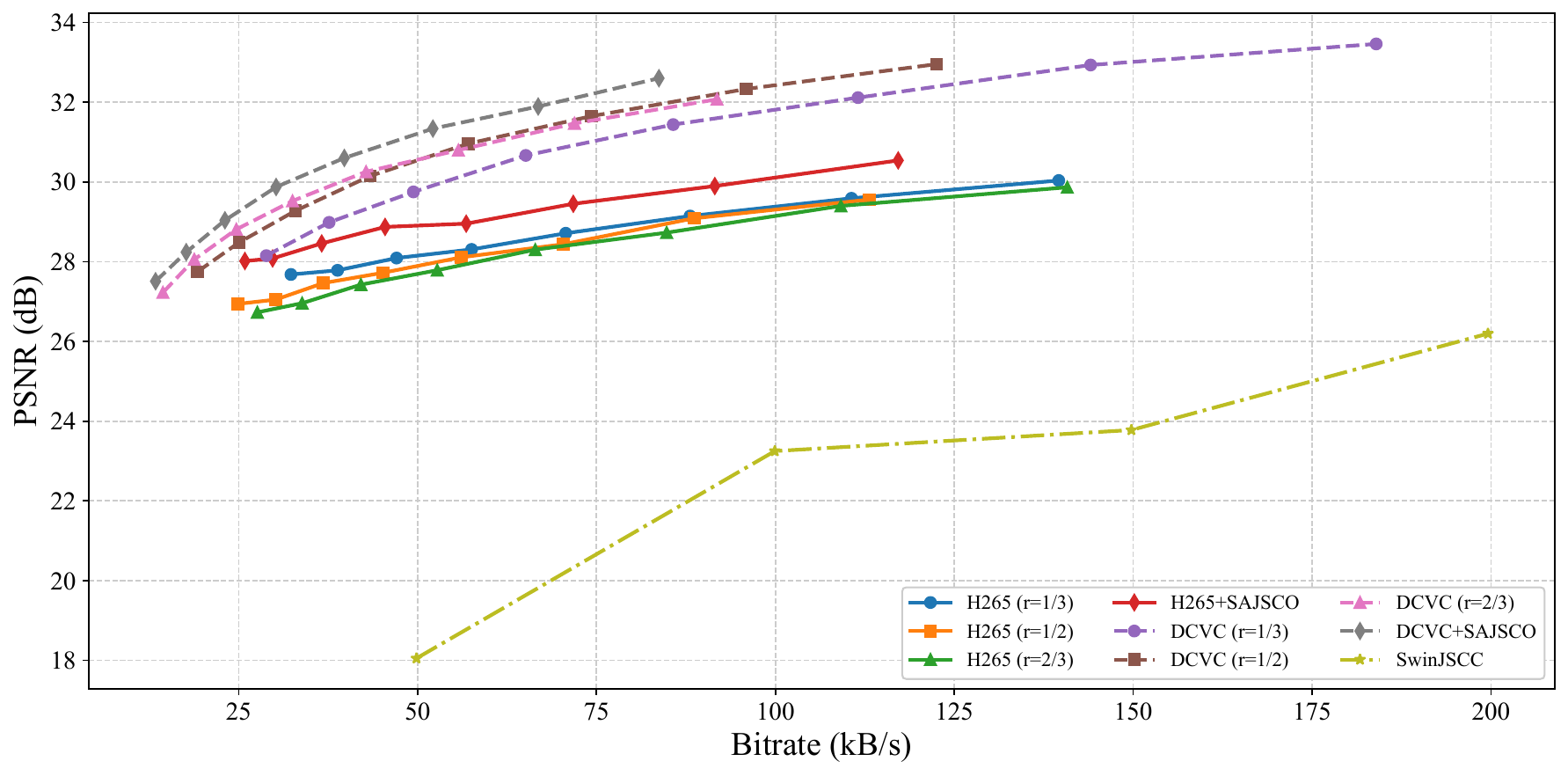}
        \label{fig:hevc_bitrate_psnr}
    }
    \hfill
    \subfloat[WPSNR versus bitrate]{
        \includegraphics[width=0.48\linewidth]{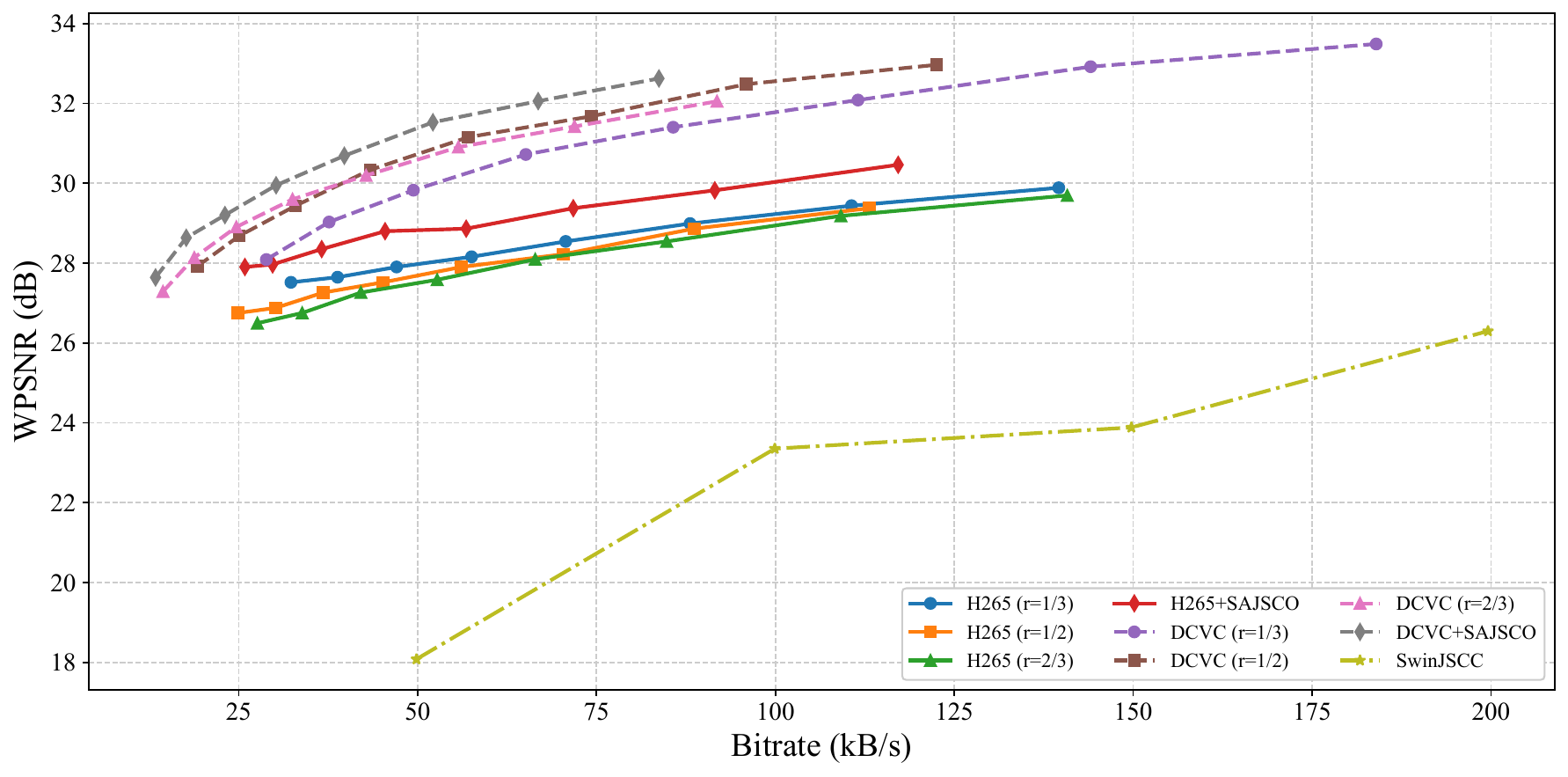}
        \label{fig:hevc_bitrate_wpsnr}
    }\\
    \subfloat[LPIPS versus bitrate]{
        \includegraphics[width=0.48\linewidth]{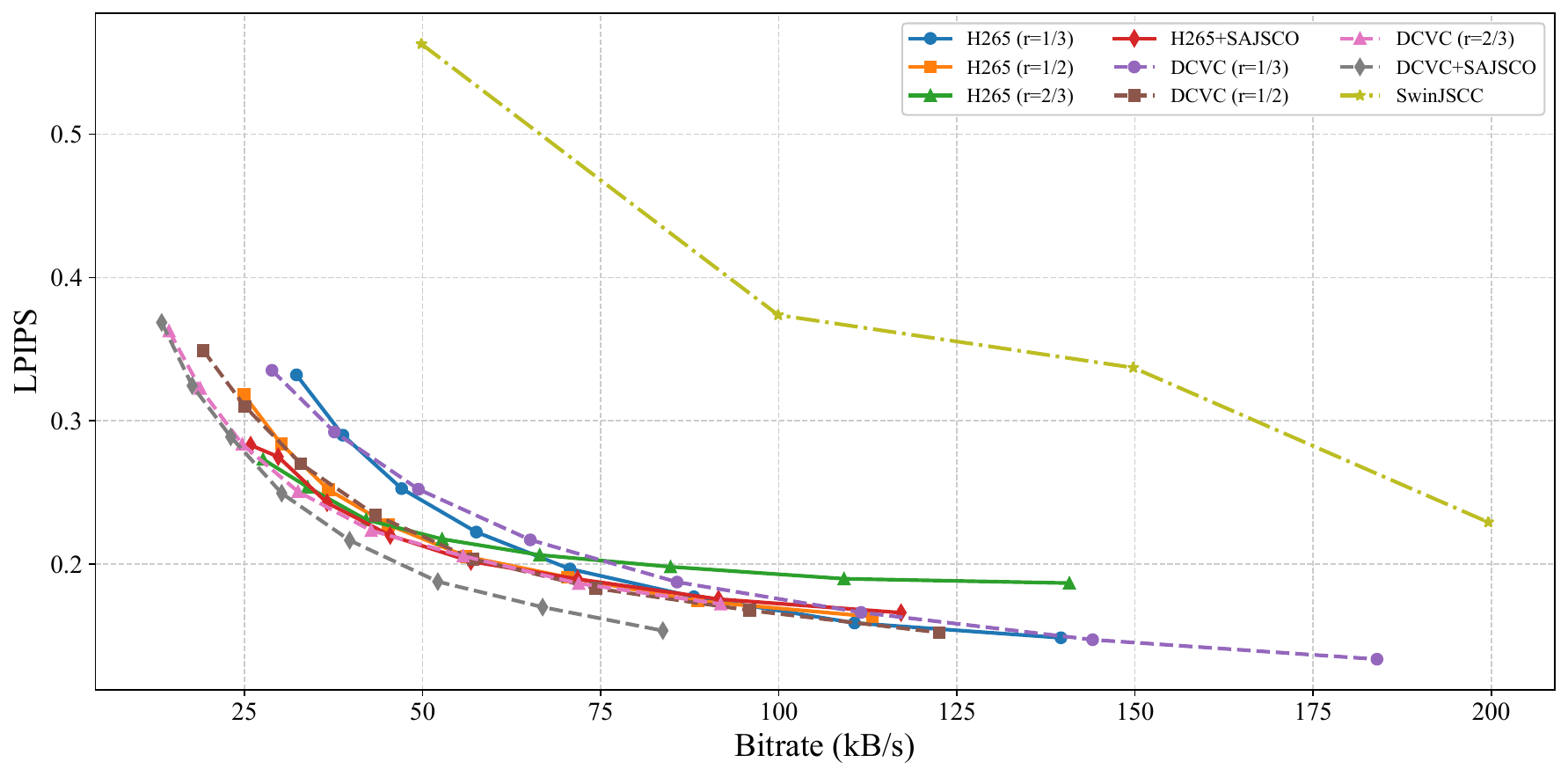}
        \label{fig:hevc_bitrate_lpips}
    }
    \hfill
    \subfloat[WLPIPS versus bitrate]{
        \includegraphics[width=0.48\linewidth]{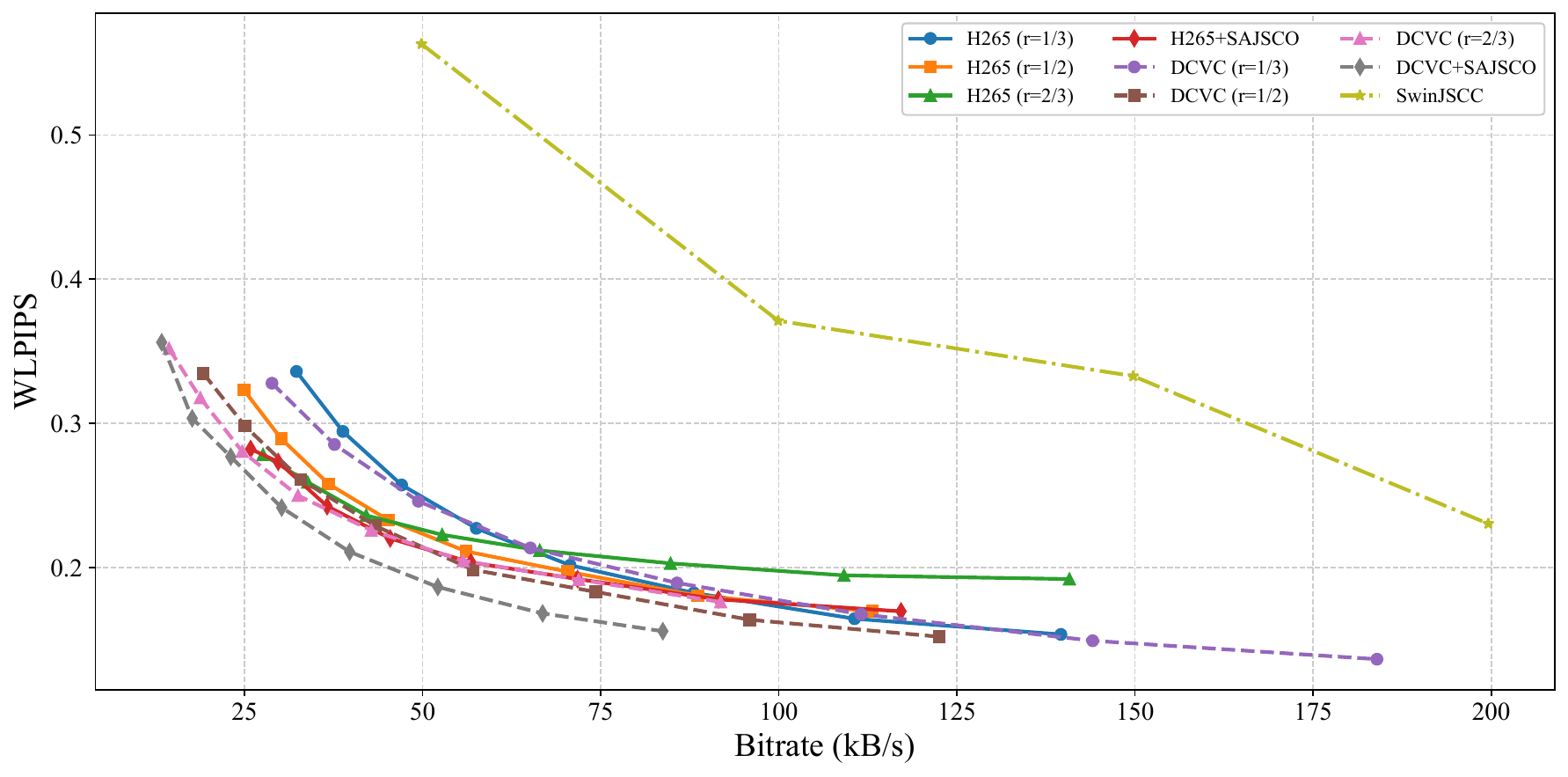}
        \label{fig:hevc_bitrate_wlpips}
    }
    \caption{Performance comparison under the RadioML CSI setting on the HEVC testset.}
    \label{fig:hevc_bitrate_results}
\end{figure*}

We conduct simulations on HEVC test dataset class D. As shown in Fig.~\ref{fig:hevc_bitrate_results}, nine curves are plotted in each subfigure, including the proposed SAJSCO, the non-semantic baselines and spatial-semantic baseline SwinJSCC. It can be observed that the proposed SAJSCO achieves better performance with both H.265 and DCVC-RT, while significantly outperforming the spatial-semantic baseline SwinJSCC. The proposed method achieves consistent BD-rate gains over the non-semantic baselines for both H.265 and DCVC-RT. Compared with the best-performing non-semantic baseline, the proposed SAJSCO scheme still achieves BD-rate reductions of $34.86\%$, $37.58\%$, $4.12\%$ and $9.17\%$ under H.265-based transmission in terms of PSNR, WPSNR, LPIPS, and WLPIPS, respectively.  When DCVC-RT is adopted as the video encoder, SAJSCO further reduces the BD-rate by $18.01\%$, $21.77\%$, $10.24\%$, and $16.41\%$, respectively, compared with the strongest non-semantic baseline. These gains mainly come from the ability of the proposed MPPO framework to adaptively adjust the source and channel coding parameters according to the time-varying CSI and the semantic importance of video content, whereas the non-semantic baselines use fixed coding parameters throughout the transmission process. 

\begin{figure*}[!t]
    \centering
    \includegraphics[width=\textwidth]{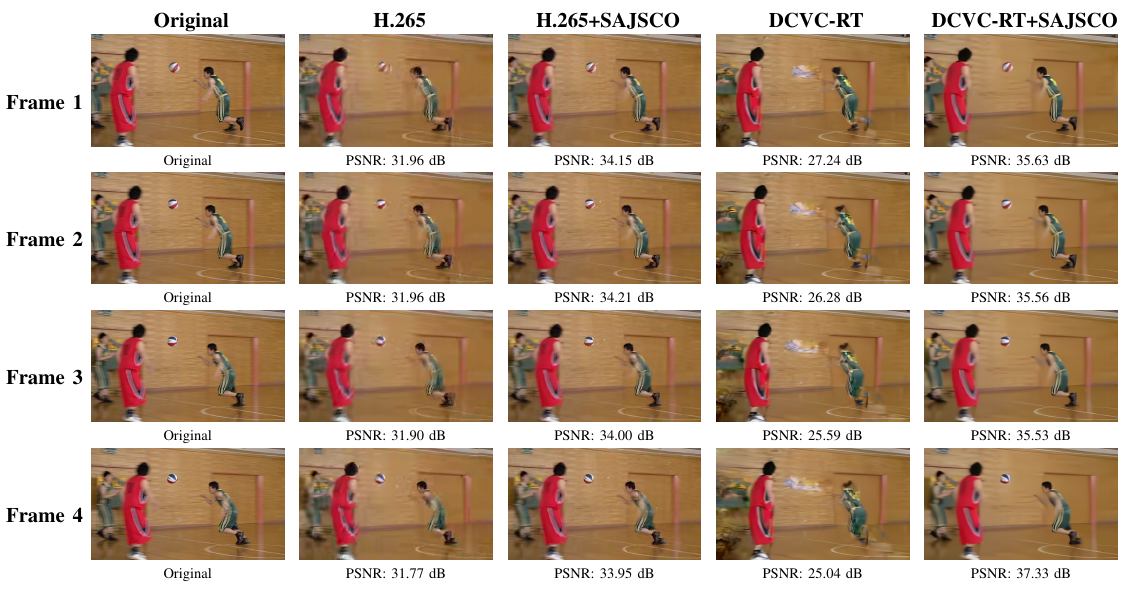}
    \caption{Visualization comparison of reconstructed video frames under different transmission schemes.}
    \label{fig:visualization_results}
\end{figure*}

\subsection{Complexity and Visualization}
Fig.~\ref{fig:visualization_results} presents the visualization results of the 61st--64th frames of the HEVC testset classD \textit{BasketballPass} video sequence, which belong to the 8th GOP under the time-varying channel condition. The corresponding SNR is $13.7~\mathrm{dB}$. Four transmission schemes are compared. Under the current bitrate setting, H.265 is combined with the best-performing LDPC coding rate of 1/3, while DCVC-RT is combined with the best-performing LDPC coding rate of 2/3. Due to the different channel coding rates, the two baselines exhibit different levels of error resilience. As shown in the figure, the reconstructed frames are noticeably degraded by channel noise in the current GOP. In particular, DCVC-RT suffers from more severe visual distortion than H.265 under this channel condition. By contrast, after introducing the proposed SAJSCO strategy, the reconstructed frames of both coding schemes exhibit greater robustness 
against channel noise, with more visual details preserved and better overall quality achieved. This further demonstrates the effectiveness and superiority of the proposed method in robust video transmission over time-varying wireless channels.

\begin{table}[t]
\centering
\caption{Comparison of Complexity}
\label{tab:complexity}
\begin{tabular}{l c c c}
\toprule
\textbf{Method} & \textbf{MACs} & \textbf{Params} & \textbf{Speed} \\
\midrule
H.265 & none & none & Enc. 42.41 fps, Dec. 25 fps \\
SAJSCO+H.265 & 318.97M & 2.23M & Enc. 35.73 fps, Dec. 25 fps \\
DCVC-RT & 385G & 20.7M & Enc. 109 fps, Dec. 90 fps \\
SAJSCO+DCVC-RT & 385.32G & 22.93M & Enc. 66.73 fps, Dec. 90 fps \\
\bottomrule
\end{tabular}
\end{table}

We tested H.265, H.265+SAJSCO, DCVC-RT and DCVC-RT+SAJSCO in terms of MACs, parameters, and encoding/decoding speed for 1080p videos, as summarized in TABLE \ref{tab:complexity}. It should be noted that H.265 is implemented with CPU-based encoding and decoding, whereas DCVC-RT is accelerated using GPU-based encoding and decoding. Compared to non-semantic baselines, the SAJSCO scheme incurs only 4.36 ms of extra latency per frame while maintaining real-time encoding capability, with a negligible increase in parameters. 

\begin{figure}[!t]
    \centering
    \subfloat[Visualization of the state space.]{
        \includegraphics[width=\linewidth]{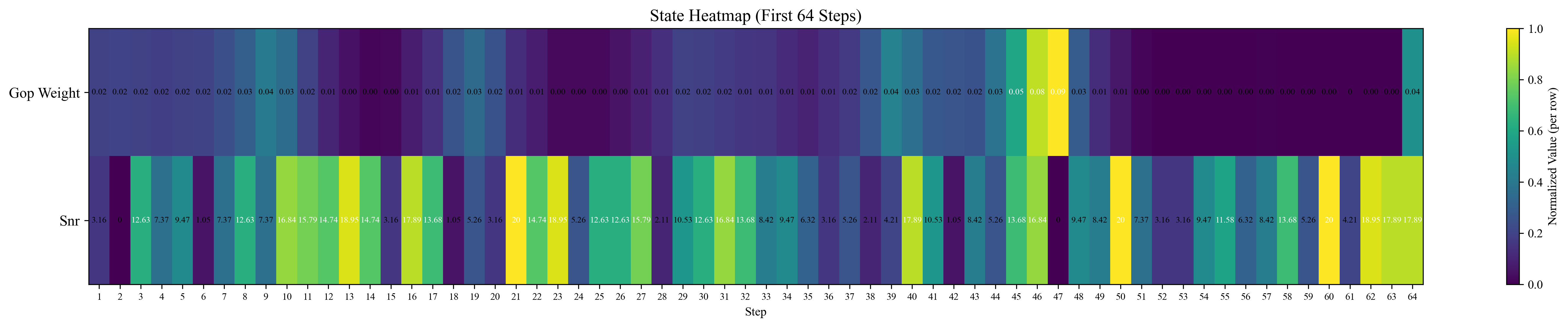}
        \label{fig:state_heatmap}
    }\\
    \subfloat[Visualization of the selected actions.]{
        \includegraphics[width=\linewidth]{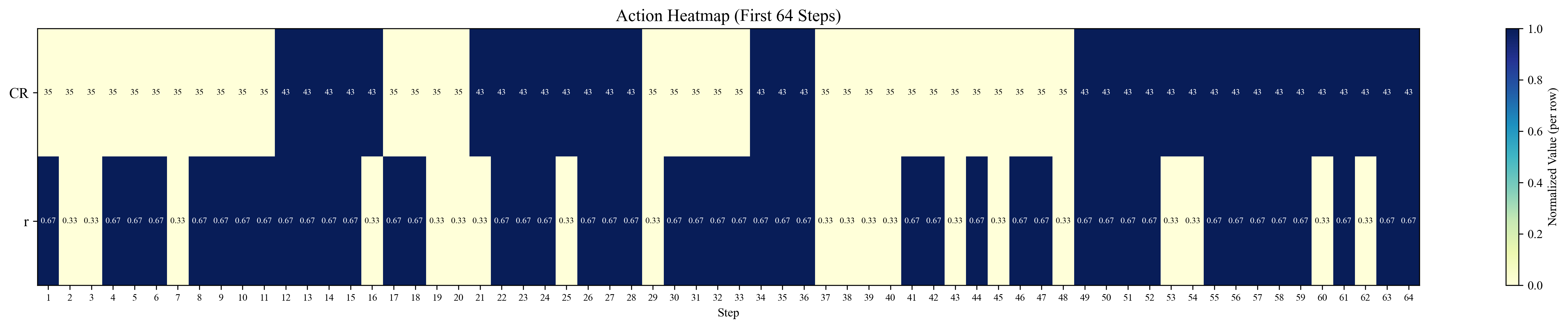}
        \label{fig:action_heatmap}
    }
    \caption{Visualization of the state space and the corresponding action selection of the proposed MPPO algorithm.}
    \label{fig:state_action_heatmap}
\end{figure}

To further illustrate the decision mechanism of the proposed MPPO algorithm, Fig.~\ref{fig:state_action_heatmap} visualizes the state space, including the CSI represented by SNR and the inter-frame semantic importance, as well as the corresponding action selection, including the source coding parameter $CR$ and the channel coding rate $r$. As shown in Fig.~\ref{fig:state_heatmap}, both the channel condition and the semantic importance vary over time, leading to a dynamically changing transmission environment. Correspondingly, Fig.~\ref{fig:action_heatmap} shows that the proposed algorithm adjusts the source and channel coding parameters adaptively. Specifically, SAJSCO adaptively selects more robust coding parameters for semantically important GOPs or unfavorable channel conditions, while adopting more efficient compression under lower semantic importance or favorable channel conditions to reduce transmission cost. This behavior demonstrates that the proposed MPPO algorithm is able to jointly exploit CSI and semantic importance information to adaptively select appropriate source-channel coding parameters, thereby achieving a better tradeoff between transmission reliability and coding efficiency.

\section{Prototype Validation}\label{sec:PROTOTYPE VALIDATION}

To further evaluate the practical performance of the proposed algorithm beyond simulation, 
we implement and validate it on a real-world communication testbed. 
The testbed is constructed using two Universal Software Radio Peripheral (USRP) devices, 
which serve as the transmitter and receiver, respectively.
The remainder of this section is organized as follows. 
We first describe the overall hardware architecture of the prototype platform, followed by a presentation and analysis of the experimental results 
obtained from the physical testbed.

\subsection{Testbed Architecture}

\begin{figure}[htbp]
    \centering
    \includegraphics[width=\columnwidth]{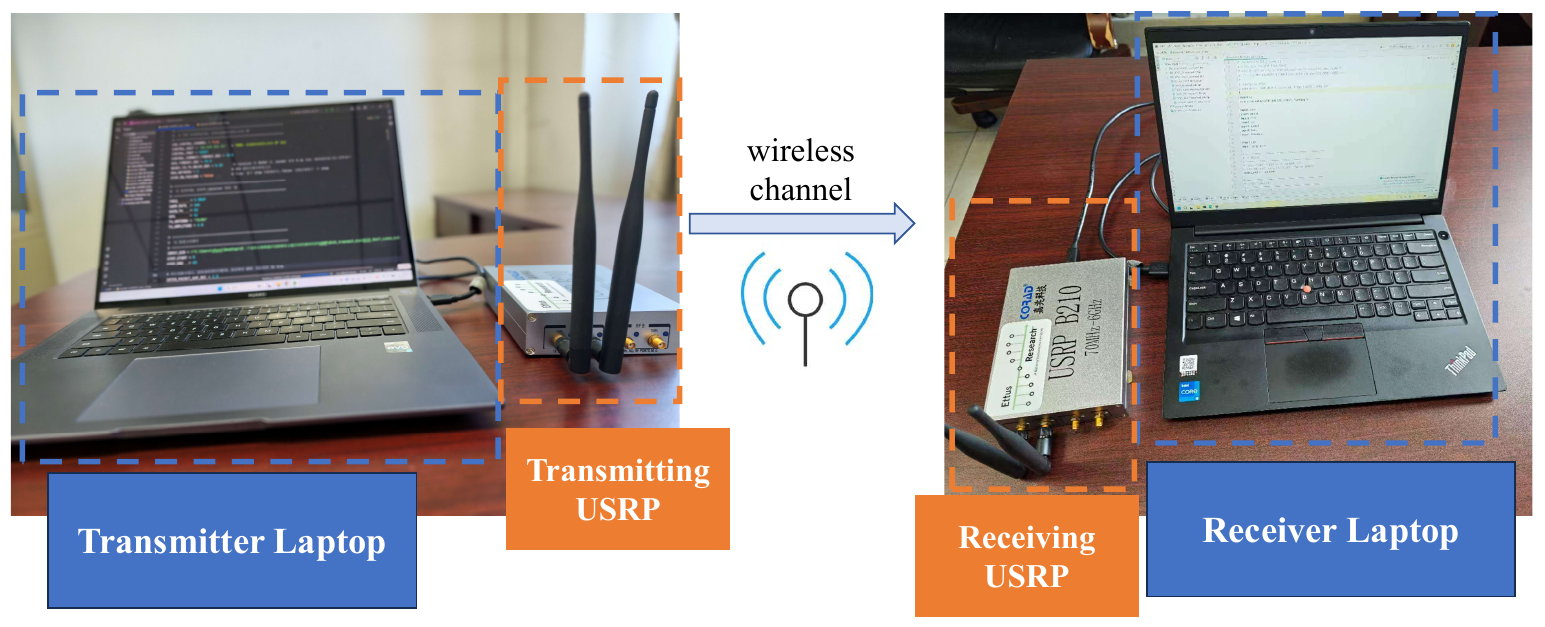}
    \caption{Hardware architecture of the proposed USRP-based communication testbed.}
    \label{fig:usrp_platform}
\end{figure}

\begin{table}[ht]
    \centering
    \caption{Experimental Parameters}
    \label{tab:usrp_params}
    \begin{tabular}{ll|ll}
        \toprule
        \textbf{Parameter} & \textbf{Value} & \textbf{Parameter} & \textbf{Value} \\
        \midrule
        Carrier Frequency       & 5.5\,GHz
            & LDPC Block Length       & 1800\,bits \\
        Sampling Rate           & 1\,MSps
            & TX Gain (Seg. 1\,\&\,4) & 50\,dB \\
        Symbol Rate             & 62.5\,kSps
            & TX Gain (Seg. 2\,\&\,3) & 60\,dB \\
        Bandwidth               & 125\,kHz
            & RX Gain                 & 56\,dB \\
        Modulation              & BPSK
            & Antenna Model           & VERT2450 \\
        Samples per Symbol      & 16
            & Antenna Placement       & Vertical \\
        Preamble Length         & 16384\,bits
            & Link Distance           & 160\,cm \\
        \bottomrule
    \end{tabular}
\end{table}

As depicted in Fig.~\ref{fig:usrp_platform}, the prototype testbed consists of two laptops and two USRP B210 software-defined 
radio (SDR) devices, where each laptop is connected to and controls one USRP unit 
via USB. The two USRPs function as the transmitter and receiver. The detailed radio frequency and hardware configuration parameters are summarized in Table~\ref{tab:usrp_params}.
On the software side, the USRP devices are controlled via Python scripts using 
the USRP Hardware Driver (UHD), which provides low-level access to 
the radio front-end for baseband signal transmission and reception.

\subsection{Experimental Results and Analysis}

The experiments employs a packet-based single-carrier BPSK waveform at a carrier frequency of 5.5\,GHz over a line-of-sight (LoS) transmission link. The transmit gain is varied across four equal-duration segments throughout the transmission: The first and last segments operate at 50\,dB, while the two 
intermediate segments operate at 60\,dB. The receive gain is maintained at a constant 56\,dB throughout all segments. This dynamic gain profile is designed to simulate the channel variations commonly encountered in practical wireless communication scenarios.

As shown in Table~\ref{tab:usrp_results}, experiments are conducted with two video encoders, namely H.265 and DCVC-RT. In both cases, SAJSCO achieves higher reconstruction quality and decoding success rate than the corresponding non-semantic baselines. Specifically, H.265+SAJSCO achieves a best PSNR of 28.905\,dB 
and a decoding success rate of 93.75\%, surpassing the best fixed-parameter H.265 baseline ($r=1/2$) by 1.448\,dB in PSNR and 4.695\% in decoding success rate. Similarly, DCVC+SAJSCO achieves a best LPIPS of 0.306, outperforming the best fixed-parameter DCVC baseline ($r=1/3$) by 0.033. H.265-based methods yield superior PSNR, WPSNR, and decoding success rate, while DCVC-RT-encoded bitstreams are more susceptible to decoding failure due to the absence of a fixed packetization structure, yet achieve more favorable perceptual quality as reflected by lower LPIPS and WLPIPS scores. These results confirm that the performance gains observed in simulation consistently transfer to real-world wireless transmission, validating the practical deployability of the proposed scheme on commodity SDR hardware.

\begin{table}[htbp]
    \centering
    \caption{Performance Comparison on the USRP Testbed}
    \label{tab:usrp_results}
    \scriptsize
    \setlength{\tabcolsep}{4pt}
    \begin{tabular}{lccccc}
        \toprule
        \textbf{Method} & \textbf{PSNR} & \textbf{WPSNR} & \textbf{LPIPS} & \textbf{WLPIPS} & \textbf{Dec. Rate} \\
        \midrule
        H.265+SAJSCO       & \textbf{28.905} & \textbf{28.922} & 0.358          & 0.351          & \textbf{93.75\%} \\
        H.265 ($r=1/3$)    & 25.358          & 25.578          & 0.518          & 0.508          & 84.38\%          \\
        H.265 ($r=1/2$)    & 27.457          & 27.514          & 0.400          & 0.403          & 89.06\%          \\
        H.265 ($r=2/3$)    & 27.262          & 27.425          & 0.527          & 0.544          & 92.19\%          \\
        DCVC+SAJSCO        & 27.741          & 27.585          & \textbf{0.306} & \textbf{0.311} & 75.00\%          \\
        DCVC ($r=1/3$)     & 26.455          & 26.084          & 0.339          & 0.351          & 70.31\%          \\
        DCVC ($r=1/2$)     & 23.917          & 23.753          & 0.394          & 0.400          & 46.88\%          \\
        DCVC ($r=2/3$)     & 26.074          & 26.266          & 0.350          & 0.347          & 60.94\%          \\
        \bottomrule
    \end{tabular}
\end{table}

\section{Conclusion}\label{sec:CONCLUSION}

This paper proposed SAJSCO, a lightweight, encoder-agnostic plug-in framework for semantic-aware joint source--channel optimization in video communication. Within this framework, we constructed a video encoding parameter optimization model by jointly considering inter-frame semantic importance and channel state information. By quantifying inter-frame semantic importance through shifted-window cosine similarity and adaptively selecting source--channel parameters via a multi-actor PPO algorithm, the proposed scheme balances reconstruction quality against transmission cost under time-varying channels. Extensive simulation results showed that the proposed scheme achieves significant performance gains over both traditional and deep video encoders, with 34.86\% and 18.01\% BD-rate reductions, respectively. Furthermore, 
the proposed scheme is implemented and validated on a real-world USRP-based 
communication testbed, where experimental results consistently confirm the performance improvements observed in simulation.


\bibliographystyle{IEEEtran}
\bibliography{mybib_0615}

\newpage

 
\vspace{11pt}




\vfill

\end{document}